\documentclass[10pt,twocolumn,letterpaper]{article}

\usepackage[pagenumbers]{cvpr} 

\usepackage[table, dvipsnames]{xcolor}
\usepackage{bm}

\definecolor{myred}{rgb}{106, 0, 11}

\usepackage{listings}

\definecolor{cvprblue}{rgb}{0.21,0.49,0.74}
\usepackage[pagebackref,breaklinks,colorlinks,citecolor=cvprblue]{hyperref}

\def\paperID{230} 
\def\confName{3DV\xspace}
\def\confYear{2025\xspace}
\begin{document}

\twocolumn[{%
\renewcommand\twocolumn[1][]{#1}%
\title{Spurfies: Sparse-view Surface Reconstruction using Local Geometry Priors}
\author{Kevin Raj$^1$ \quad Christopher Wewer$^1$ \quad Raza Yunus$^1$ \quad
Eddy Ilg$^2$ \quad Jan Eric Lenssen$^1$
\\$^1$Max Planck Institute for Informatics, Saarland Informatics Campus, Germany \\
$^2$Saarland University, Saarland Informatrics Campus, Germany
\\
{\tt\small{\{kraj,jlenssen\}}@mpi-inf.mpg.de}
}

\maketitle
\begin{center}
\centering
\captionsetup{type=figure}
\vspace{-0.2cm}
\includegraphics[width=1\textwidth]{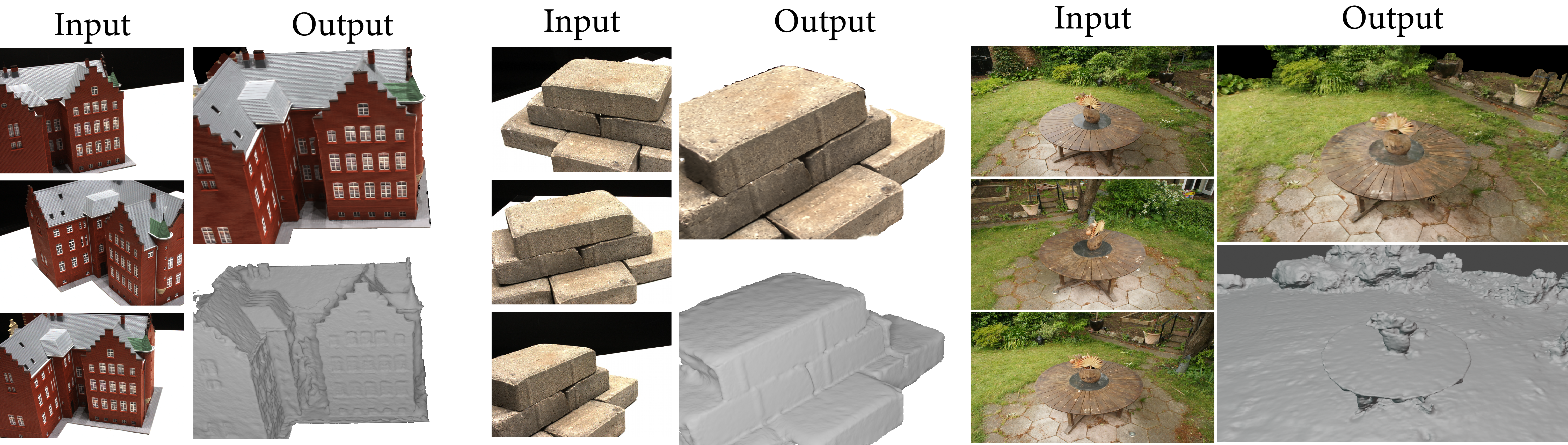}
\captionof{figure}{Spurfies leverages synthetic data to learn local surface priors for surface reconstruction from few images. Our model significantly outperforms previous methods and can be applied to both bounded (DTU dataset, left) and unbounded scenes (Mip-NeRF360, right).}
\label{fig:teaser}
\end{center}%
}]

\begin{abstract}
\vspace{-0.2cm}
We introduce Spurfies\footnotemark, a novel method for \underline{sp}arse-view s\underline{urf}ace reconstruct\underline{i}on that disentangles appearance and geometry information to utilize local g\underline{e}ometry prior\underline{s} trained on synthetic data.
Recent research heavily focuses on 3D reconstruction using dense multi-view setups, typically requiring hundreds of images. However, these methods often struggle with few-view scenarios. Existing sparse-view reconstruction techniques often rely on multi-view stereo networks that need to learn joint priors for geometry and appearance from a large amount of data. In contrast, we introduce a neural point representation that disentangles geometry and appearance to train a local geometry prior using a subset of the synthetic ShapeNet dataset only. During inference, we utilize this surface prior as additional constraint for surface and appearance reconstruction from sparse input views via differentiable volume rendering, restricting the space of possible solutions. We validate the effectiveness of our method on the DTU dataset and demonstrate that it outperforms previous state of the art by $35\%$ in surface quality while achieving competitive novel view synthesis quality. Moreover, in contrast to previous works, our method can be applied to larger, unbounded scenes, such as Mip-NeRF360. 
\end{abstract}
\section{Introduction}
\label{sec:intro}
3D surface reconstruction just from a small number of 2D input images has been an important goal in computer vision in the recent years. While we entered an era in which we can get high-quality reconstructions from a large number of carefully captured images through recent advances in SDF-based neural fields~\cite{neus, neus2} or surfels in the form of 2D Gaussians~\cite{2dgs}, we still lack the capabilities to perform this task from a sparse set of views. The key challenge in this setup is the under-constrained nature of the optimization problem at hand: there are simply too many potential combinations of geometry and appearance that would satisfy the sparse set of given image observations.
\footnotetext{\href{https://geometric-rl.mpi-inf.mpg.de/spurfies/}{geometric-rl.mpi-inf.mpg.de/spurfies/}}

The usual way to solve such under-constrained tasks is to introduce regularization or priors into the formulation. 
Typical approaches include feature consistency~\cite{neusurf} and cues from monocular depth~\cite{monosdf}, multi-view stereo~\cite{neuris, s-volsdf}, or structure from motion~\cite{geoneus}.

More recently, learned priors have been introduced, e.g. from pre-trained 2D diffusion models~\cite{Wu2024CVPR}. 
They have been shown to clearly improve reconstruction quality from a few observations, being able to hallucinate missing appearance details. However, a clear limitation of diffusion-based approaches is that they require large amounts of training data in order to generalize to the diverse domain of appearances that 3D scenes can have.

In this work, we make the key observation that the space of local surface geometry is much less diverse than that of surface appearance and make the following three key assumptions: (1) we argue that much less training data is needed to learn a useful surface geometry prior than to learn a prior about appearance, (2) that synthetic data can serve well as training data for geometry as the domain gap between real and synthetic data for geometry is smaller, and (3) that a geometry prior already serves as a useful constraint to appearance reconstruction from only a few observations. To verify, we design an architecture that explicitly disentangles geometry from appearance modeling, while still jointly optimizing both during the reconstruction process. With this architecture, we are able to pre-train the geometry part of our network on synthetic data and enable it to learn the space of common surface structures. During test-time optimization, the pre-trained geometry branch then provides regularization for the joint SDF and appearance reconstruction of the scene.

As neural field representation we opt for a point-based neural field~\cite{pointnerf}, with points initialized by DUSt3R~\cite{dust3r}, which we extend to model signed distance functions (SDFs) instead of density. On the neural point cloud we store separate features for SDFs and appearance in local neighborhoods. The representation is rendered via volume rendering by aggregating and processing neighboring features for sample points along the ray.
Signed distance and radiance are predicted by a pre-trained geometry and a per-scene appearance decoder, respectively.
 
We show that Spurfies 
\begin{itemize}
\item outperforms previous methods in sparse-view surface reconstruction quality by a large margin,
\item reaches state-of-the-art novel view synthesis quality, and
\item can be applied to larger, unbounded scenes,
\end{itemize}
by relying on recent advances in point reconstruction~\cite{dust3r} and a local surface prior trained purely on synthetic data.

\section{Related Work}
\subsection{Surface Reconstruction from Images}
\paragraph{Dense Views}
Since NeRF~\cite{nerf} revolutionized inverse graphics by parameterizing implicit radiance fields as neural networks trained via differentiable volume rendering, there have been many approaches to directly reconstruct 3D surfaces in a similar manner.
NeuS~\cite{neus} and VolSDF~\cite{volsdf} reformulate the density of radiance fields based on a signed distance function to implicitly learn a surface representation via inverse rendering.
Follow-up works like MonoSDF~\cite{monosdf}, Geo-Neus~\cite{geoneus}, RegSDF~\cite{regsdf}, and NeuRIS~\cite{neuris} leverage depth and normal estimation, sparse geometry from structure from motion (SfM), and photometric consistency constraints in multi-view stereo for additional regularization.
Another line of research~\cite{instantnsr,bakedsdf,neuralangelo,voxurf,neus2} focuses on accelerating training and rendering of neural surface representations by incorporating explicit data structures like multi-resolution hash grids~\cite{instantngp}.
To this end, following the breakthrough of fully explicit 3D Gaussian Splatting~\cite{3dgs}, SuGaR~\cite{sugar}, 2D GS~\cite{2dgs}, and DN-Splatter~\cite{dnsplatter} try to tame Gaussians for accurate mesh reconstruction.
While high-fidelity surface reconstruction is now possible with training in the matter of minutes, all these approaches assume dense captures of the scene in the form of a large number of input images.

\paragraph{Sparse Views}
To reduce the number of input views and therefore enable applications to more real-world settings, geometric priors can further constrain the space of possible reconstructions. Approaches for generalizable surface reconstruction~\cite{sparseneus, volrecon, retr, c2f2neus, s-volsdf} learn such priors on additional training scenes.
Specifically, SparseNeuS~\cite{sparseneus} fuses 2D features into a cascade of coarse-to-fine voxel grids for end-to-end SDF prediction from sparse input views. These predictions can be used as initialization for per-scene finetuning. VolRecon~\cite{volrecon} and ReTR~\cite{retr} leverage transformer architectures for rendering conditioned on similar fused voxel features. C2F2NeUS~\cite{c2f2neus} replaces imprecise, regular voxel grids with view-based cascade cost frustums.
S-VolSDF~\cite{s-volsdf} combines VolSDF~\cite{volsdf} with CasMVSNet~\cite{casmvsnet} by supervising the former with probability volumes from the latter, which in turn produces depth hypotheses used for sampling in a finer MVS stage.
All of these approaches train a surface prior on image data, limiting generalization to scenes close to the training distribution.

NeuSurf~\cite{neusurf} proposes a two-stage approach by first training an unsigned distance field given sparse point clouds from SfM~\cite{cap-udf}, which is used as a coarse geometric constraint for fitting a fine signed distance field in the second stage.
However, it relies on global representations resulting in limited efficiency, scalability, and level of detail.
In contrast to that, our method trains a local geometric prior on general, synthetic, texture-less meshes, requiring little data for generalization.
We achieve this by decomposing our architecture into geometry and appearance branches build on a point-based representation. Given the recent breakthrough of DUSt3R~\cite{dust3r} for stereo 3D point cloud reconstruction via large-scale (pre-)training of transformer architectures, point-based representations render an elegant solution to obtain watertight surfaces. 

\subsection{Point-based 3D Representations}
Point-based 3D representations~\cite{pbgraphics, pbgraphics+, pointnerf, 3dgs} equip point clouds obtained from \mbox{RGB-D} sensors, MVS, or SfM pipelines with additional learnable parameters.
Point-NeRF~\cite{pointnerf} decodes these features with small MLPs to represent a radiance field fitted on dense input views.
Subsequent works employ similar representations for relighting and deformation~\cite{spidr}, single-view reconstruction~\cite{simnp}, and object generation~\cite{npcd}.
3D GS~\cite{3dgs} and follow-ups for surface~\cite{sugar, 2dgs, dnsplatter}, generalizable~\cite{pixelsplat, latentsplat, mvsplat}, or non-rigid~\cite{dynamicgaussians, npg, 4dgs} reconstruction use optimizable Gaussian parameters and spherical harmonic coefficients for view-dependent RGB, enabling much faster rasterization-based rendering without any neural components.
However, the explicit nature is hardly compatible with learning of neural priors without loosing the efficiency advantages.
Therefore, our architecture is based on neural point clouds with individual geometry and appearance decoding branches for data-efficient pre-training of a local surface prior only.
\section {Prerequisites}
\label{sec:prereq}

\newcommand{\vx}{\mathbf{x}}

Given a sparse set of $N = 3$ RGB images 
$\mathcal{I} = \{\mathbf{I}_i \,|\, i \in 1, \dots, N\}$, known camera poses 
$\mathcal{V} = \{\mathbf{v}_i \,|\, i \in 1, \dots, N\}$ and a point-cloud 
$\mathcal{P} = \{\mathbf{p}_i \,|\, i \in 1, \dots, M \}$  that we obtain from DUSt3R~\cite{dust3r}, 
we aim to reconstruct a surface $\mathcal{S}$. In the following, we denote all predictions from our model with the $\,\hat{}$-symbol. We choose a continuous implicit surface representation using a signed distance field $\hat{s}(\vx)$ and use the approach from VolSDF~\cite{volsdf} by attaching a volumetric radiance field with density $\hat{\sigma}(\vx)$ and radiance $\hat{r}(\vx)$ to leverage its graceful optimization properties via backpropagation and gradient descent. The density is derived from the signed distance via: 
\begin{equation}
    \hat{\sigma}(\mathbf{x}) = 
\begin{cases} 
\frac{1}{2} \exp\left(\frac{\mathrm{\hat{s}(\vx)}}{\beta}\right) \cdot \alpha & \text{if } \mathrm{\hat{s}(\vx)} \leq 0 \\
\left(1 - \frac{1}{2} \exp\left(-\frac{\hat{s}(\vx)}{\beta}\right)\right) \cdot \alpha & \text{if } \mathrm{\hat{s}(\vx)} > 0 \mathrm{\,,}
\end{cases}
\end{equation}
where $\alpha$ and $\beta$ are learnable parameters that control the sharpness of the surface. For a given ray $\vx(t) = \mathbf{o} + t \cdot \mathbf{d}$, we use importance sampling to sample points with intervals $\Delta_i$ to obtain density $\hat{\sigma}_i$ and radiance $\hat{r}_i$.
We then compute the final pixel colors $\hat{C}_{u,v}$ for pixel $(u,v)$ via transmittance computation and ray accumulation from NeRF~\cite{nerf, volsdf}.

\begin{figure*}[!t]
    \centering
    \includegraphics[width=\textwidth, keepaspectratio]{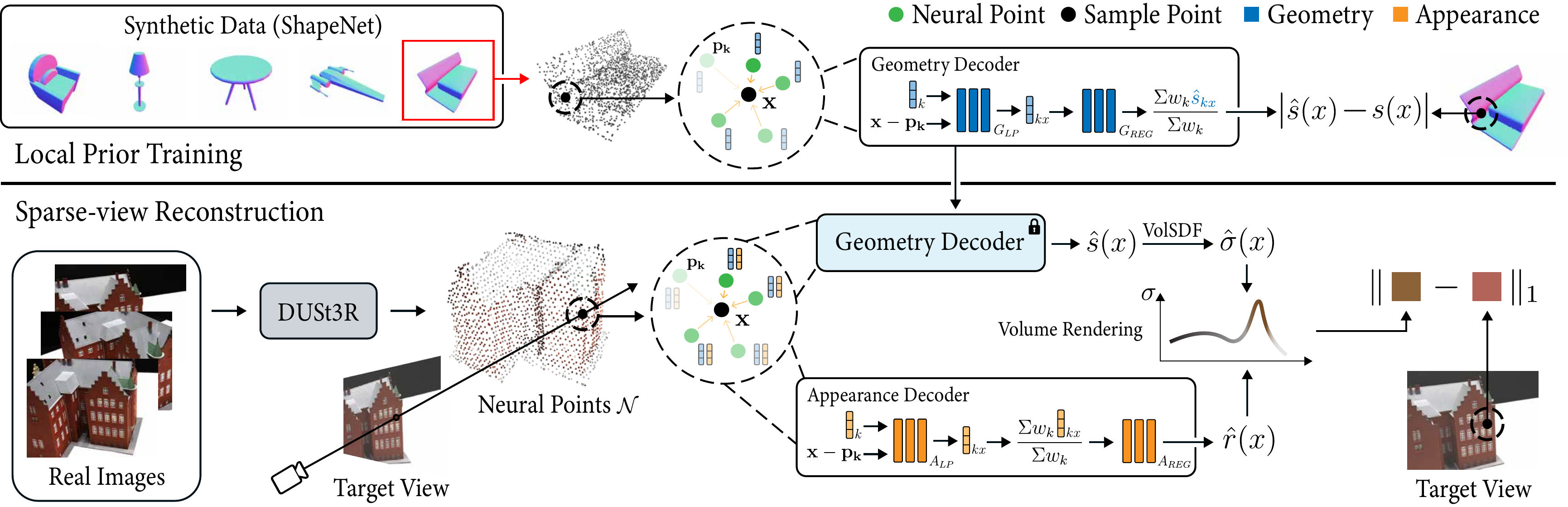}
    \caption{\textbf{Method overview:} 1) \textbf{Preprocess:} given a sparse set of input views, we make use of DUSt3R~\cite{dust3r} to predict points $\mathcal{P}$. \textbf{Representation:} The points serve as basis for a neural point representation that stores disentangled features $\mathbf{f}^a$, $\mathbf{f}^g$ for geometry and appearance on each point. \textbf{Local Prior} (top): We learn a local geometry prior $G_\textnormal{LP}$ \& $G_\textnormal{REG}$ over a subset of shapes from the synthetic ShapeNet dataset~\cite{shapenet} by optimizing to predict ground truth SDF. 3) \textbf{Spurfies} (bottom): We make use of the prior for surface reconstruction from sparse images, only optimizing the latent codes $\mathbf{f}^a, \mathbf{f}^g$ and the color MLPs $A_\textnormal{LP}$ \& $A_\textnormal{REG}$ to reconstruct images via volume rendering.}
    \label{fig:main}
\end{figure*}

\section{Spurfies}

As the combination of signed distance fields with volumetric rendering introduced in the last paragraph has shown to provide high-quality surface reconstructions from many observations, the question is how to enable sparse-view reconstruction by representing $\hat{s}(\vx)$ and $\hat{r}(\vx)$ with a data prior that was obtained from a training dataset. We introduce our technique that we name \emph{Spurfies} (Figure~\ref{fig:main}). It extends volumetric rendering by two key components: 
\textbf{\mbox{(1) A neural} point representation} (Section \ref{sec:representation}), where each point $\mathbf{p}_i$ is associated with geometry and appearance latent codes $\mathbf{f}_i = (\mathbf{f}^g_i, \mathbf{f}^a_i)$ that enable local processing and effective disentanglement of both modalities. 
\textbf{(2) A local geometry prior} (Section~\ref{sec:lp}) to mitigate the shape-radiance ambiguity inherent in sparse-view reconstruction~\cite{nerf++}. We argue that there is less variety in geometry than in appearance such that disentanglement in combination with a strong geometry prior is key. Unlike previous works that rely on multi-view stereo data during training, it is sufficient to obtain such a local prior from a synthetic object dataset (ShapeNet~\cite{shapenet}). 

\subsection{Neural Point Representation}
\label{sec:representation}
We use a distributed representation that is applicable to whole scenes by attaching disentangled latent codes $\mathbf{f}^g_i, \mathbf{f}^a_i$ to the points $\mathbf{p}_i$ to form a neural point cloud. To regress $\hat{s}(\vx)$ and $\hat{r}(\vx)$ at a query location $\vx$, we use $K$ neural points $\mathbf{p}_k$ within a local neighborhood of radius $R$.

\paragraph{Local Processing.} 

We map the point features to the relative query location: 
\begin{align}
    \mathbf{f}^g_{kx} &= \mathit{G}_{\mathrm{LP}}(\mathbf{f}^g_k, \vx - \mathbf{p}_k) \mathrm{\,,} \\
    \mathbf{f}^a_{kx} &= \mathit{A}_{\mathrm{LP}}(\mathbf{f}^a_k, \vx - \mathbf{p}_k) \mathrm{\,,} 
\end{align}
where $\mathit{G}_{\mathrm{LP}}$ and $\mathit{A}_{\mathrm{LP}}$ are two small MLPs. The resulting intermediate latent codes $\mathbf{f}^g_{kx}$ and $\mathbf{f}^a_{kx}$ encode local geometry and appearance. The use of relative positions $\vx - \mathbf{p}_k$ makes our architecture local and translation invariant, a property that allows application on larger scenes~\cite{pointnerf}. 

\paragraph{Signed Distance Regression.} 

We then directly map the geometry features to signed distance values and interpolate them at the query location via: 
\begin{equation}
   \hat{s}(\vx)  = \frac{\sum_{k=1}^{K} w_k \cdot \hat{s}_{kx}}{\sum_{k=1}^{K} w_k}, \quad \hat{s}_{kx} = \mathit{G}_{\mathrm{REG}}(\mathbf{f}^g_{kx}), \\
\end{equation}
where $\mathit{G}_{\mathrm{REG}}$ is another MLP and the weighting function $w_k = e^{-\lambda \|\mathbf{x} - \mathbf{p}_k\|_2^2}$ is modelled using radial basis functions (RBFs). We use Gaussians as the distance kernel with variance $\lambda$, which controls the influence of the neighboring neural points $\mathbf{p}_k$. In addition, the RBF weights are bounded between $[0,1]$ providing stability during the training process. 

\paragraph{Radiance Regression.}

For radiance prediction, we first interpolate the local scene appearance encoded by neighboring points and subsequently regress the radiance with a small MLP $\mathit{A}_{\mathrm{REG}}$:
\begin{equation}
    \quad \hat{r}(\vx, \mathbf{d}) = \mathit{A}_{\mathrm{REG}}(\mathbf{f}^a_x, \mathbf{d}), \quad \mathbf{f}^a_x = \frac{\sum_{k=1}^{K} w_k \cdot \mathbf{f}^a_{kx}}{\sum_{k=1}^{K} w_k} \mathrm{\,,}
\end{equation}
with view direction $\mathbf{d}$, following previous work~\cite{pointnerf}.
The interpolation weights $w_k$ are computed using the same RBF kernel as in the SDF regression, ensuring smooth blending of local appearance information. Finally, $\hat{s}(\vx)$ and $\hat{r}(\vx)$ are rendered via volumetric rendering (c.f. Sec.~\ref{sec:prereq}).

\subsection{Local Priors}
\label{sec:lp}

To address the shape-radiance ambiguity~\cite{nerf++} inherent in sparse-view settings, we propose a novel approach that diverges from traditional multi-view stereo (MVS) based methods. While previous sparse-view reconstruction techniques~\cite{sparseneus, volrecon} often rely on MVS networks with cost volumes and use 3D CNNs to regress SDF and color, we leverage the disentangled nature of our proposed neural point representation. Notably, in our approach the geometry and appearance branches are completely separate. This enables us to learn generalizable local geometry priors in the networks $\mathit{G}_{\mathrm{LP}}$ and $\mathit{G}_\mathrm{REG}$ on a synthetic dataset with available ground-truth geometry. 
During inference, we keep the geometry networks frozen and only optimize the geometry latent codes together with the full appearance branch.

\vspace{-0.2cm}
\subsubsection{Training}

To train the geometry network effectively, we require a set of query points \( \mathcal{X} = \left\{\mathbf{x}_i\right\}_{i=1}^{S} \) with corresponding ground-truth signed distances $s(\mathbf{x}_i$). 
 
For each training mesh, we also sample a neural point-cloud \(
\mathcal{N} = \{(\mathbf{p}_i, \mathbf{f}^g_{i})\}_{i=1}^{\mathrm{M}}
\). 
It consists of global point coordinates and randomly initialized geometry latent codes for each point.

We follow the procedure outlined in ~\cite{deepsdf, deepls} to sample the query points and evaluate the ground-truth SDF. To enhance the network's robustness to noise, we add Gaussian noise to the neural point locations with variance 0.005.
Our training dataset is a small subset of ShapeNet~\cite{shapenet}, consisting of five object classes: sofas, chairs, planes, tables, and lamps. The local prior is trained on a total of 50 objects, with 10 objects randomly selected from each class.

This training approach ensures that our geometry network learns a robust and generalizable representation of local shapes using SDF ground truth as opposed to using incomplete depth maps or unreliable inverse renderings, which require much more diverse training data and are significantly more expensive. A key contribution is to show that a small and accurate synthetic dataset is sufficient to learn a prior that generalizes to general, large scale scenes. 

\vspace{-0.2cm}
\subsubsection{Loss Functions}

We train the local prior using the following loss functions.

\vspace{-0.3cm}
\paragraph{SDF Loss.} We take the absolute difference between the ground-truth ${s(\mathbf{x}_i)}$ and the predicted SDF value ${\hat{s}(\mathbf{x}_i)}$~\cite{instantngp}. We further weight the loss closer to the surface by scaling with the inverse of the ground-truth distance and add a small number to prevent division by zero: 
\begin{equation}
    \mathcal{L}_{\mathrm{SDF}} = 
    \sum_{i=1}^S
    \frac{\left| s(\mathbf{x}_i) - \hat{s}(\mathbf{x}_i) \right|}{\left| s(\mathbf{x}_i) \right| + \epsilon}.
\end{equation}

\vspace{-0.3cm}
\paragraph{Eikonal Loss.} We use the Eikonal loss to regularize the Signed Distance Function (SDF)~\cite{eik} and ensure the  properties of a true distance field: 

\begin{equation}
    \mathcal{L}_{\mathrm{Eik}} = \sum_{i=1}^S \left( \|\nabla \hat{s}(\mathbf{x}_i)\|_2 - 1 \right)^2.
\end{equation}

\begin{table*}[!t]
  \centering
  \resizebox{\textwidth}{!}{
\begin{tabular}{lcccccccccccc}
\toprule
\toprule
\textbf{Scan ID}          & \textbf{21} &\textbf{24} & \textbf{34} & \textbf{37} & \textbf{38} & \textbf{40} & \textbf{82} & \textbf{106} & \textbf{110} & \textbf{114} & \textbf{118} & \textbf{Mean CD} \\
\midrule
Points2Surf~\cite{points2surf} & 3.73 & 2.85 & 2.55 & 5.13 & 3.85 & 2.41 & 2.30 & 3.95 & 3.33 & 2.37 & 2.84 & 3.21 \\
DUSt3R~\cite{dust3r} + Poiss.~\cite{poisson} & 3.31 & 2.12 & 2.00 & 4.25 & 3.28 & 2.59 & 2.48 & 4.28 & 3.85 & 2.68 & 3.32 & 3.11 \\
CAP-UDF~\cite{cap-udf} & 2.72 & 1.53 & 1.45 & 4.05 & 2.78 & 1.81 & 4.22 & 3.51 & 3.83 & 2.24 & 3.65 & 2.89 \\
\midrule
NeuS~\cite{neus} & 4.52 & 3.33 & 3.03 & 4.77 & 1.87&  4.35 & 1.89 & 4.18 & 5.46 &  1.09 & 2.40 & 3.36 \\
VolSDF~\cite{volsdf}     &  4.54   &  2.61  &  1.51  & 4.05   & 1.27   &  3.58  &  3.48 &  2.62   &  2.79   &   0.52  &   1.10  &   2.56   \\
\midrule
SuGar~\cite{sugar}      &  2.71  &  2.04  &  2.14  &  4.01  &  2.90  &  2.45  &  4.68  &  3.82   &   3.28  &  2.44   &   2.66  &  3.01    \\
\midrule
$\text{SparseNeus}_{ft}$~\cite{sparseneus} &  3.73  & 4.48  &  3.28  &  5.21  &  3.29  &  4.21  & 3.30   &  2.73   &  3.39   & 1.40    &   2.46  &   3.41   \\
VolRecon~\cite{volrecon}   &  3.05  &  3.30  &  2.27  &  4.36  &  2.51  &  3.24  &  3.30  &  3.10   &  3.58   &  1.86   &  3.68   &  3.11   \\
S-VolSDF~\cite{s-volsdf}   &  3.18  &  2.95  &  2.19  &  3.40  &  2.30  & 2.69   &  2.69  &  1.60   & 1.48   &  1.21   &  1.16   &   2.26   \\
NeuSurf~\cite{neusurf}   &  3.22  &  2.42  &  1.38  &  2.61  &  1.72  & 3.46   &  2.68  &  1.44   & 2.42   &  \textbf{0.61}   &  \textbf{0.87}   &   2.08   \\
\midrule
\rowcolor{lightgray}
\textbf{Ours}       &  \textbf{2.36}  &  \textbf{1.12}  &  \textbf{0.83}  &  \textbf{2.39}  &  \textbf{1.14}  &  \textbf{1.55}  &  \textbf{1.67}  &  \textbf{1.26}   &  \textbf{1.14}   &   \textbf{0.61}  &  0.94   &   \textbf{1.36}   \\
\bottomrule
\bottomrule
\end{tabular}
}
\caption{\textbf{Quantitative mesh reconstruction comparison} based on Chamfer Distance (mm) $\downarrow$ on the DTU dataset. On average, we improve on the previous best method by \textbf{35}\% (\textbf{1.36} vs. 2.08 CD). We also outperform various baselines that work on points only.}
  \label{tab:CD}
\end{table*}

\vspace{-0.3cm}
\paragraph{Total Variation Loss.} We find that regularizing the geometry latent codes, such that nearby neural points are closer in latent space, is beneficial for surface reconstruction. Thus, we introduce a total variation loss: 
\begin{equation}
    \mathcal{L}_{\mathrm{TV}} =  \sum_{i=1}^{M} \sum_{k \in \mathcal{K}(i)} \frac{\|\mathbf{f}^g_i - \mathbf{f}^g_k\|_1}{\|\mathbf{p}_i - \mathbf{p}_k\|_2} \mathrm{\,,}
\end{equation}
where $\mathcal{K}(i)$ is the local neighborhood of neural points around the point with index $i$.
This results in smooth changes to the regressed local surface (c.f. Sec.~\ref{sec:ablations}). 

\vspace{-0.2cm}
\paragraph{Training Objective.}
In total, we find the optimal parameters of $G_\mathrm{LP}$, $G_\mathrm{REG}$, and $\mathbf{f}^g_i$ by minimizing:
\begin{equation}
    \mathcal{L}_{\mathrm{Prior}} = \mathcal{L}_{\mathrm{SDF}} + \lambda_{\mathrm{TV}} \cdot \mathcal{L}_{\mathrm{TV}} + \lambda_{\mathrm{Eik}} \cdot \mathcal{L}_{\mathrm{Eik}} \mathrm{\,,}
\end{equation}
where the factors $\lambda$ weigh the individual loss terms.

\subsection{Sparse-view Reconstruction}
\label{sec:svr}

We incorporate the knowledge learned from the local prior to the volume rendering pipeline. We achieve this by keeping the geometry network fixed and only optimizing latent codes and the appearance network using the following established rendering losses for sparse-view reconstruction. For details about loss functions from previous works and the individual weights we refer to the appendix.

\vspace{-0.2cm}
\paragraph{Rendering Loss.}
To ensure that our reconstructed model accurately reproduces the input images, we employ a rendering loss. It is defined as the $L_1$ distance between the rendered and ground-truth images:
\[
\mathcal{L}_{\mathrm{Ren}} = \sum_{i,u,v\in \mathcal{I}} \| \hat{C}_{i,u,v} - C_{i,u,v} \|_1 \mathrm{\,,}
\]
where $\hat{C}_{i,u,v}$ is the pixel color of pixel $(u,v)$ of image $i$, and $C_{i,u,v}$ the respective ground truth value.

\vspace{-0.2cm}
\paragraph{Feature Consistency Loss.}

We use a feature consistency loss $\mathcal{L}_{\mathrm{FC}}$ from NeuSurf~\cite{neusurf}. 
It minimizes the differences in VisMVSNet~\cite{vismvsnet} features between the projections of surface points onto different views. This encourages consistency across views and support reconstructing finer surface details.

This loss minimizes the differences in feature space between the projections of each surface point onto different views. By doing so, we encourage consistency across views and support reconstructing finer surface details.

\begin{table}[t]
\centering
\begin{tabular}{l|ccc}
\toprule
\toprule
\multicolumn{1}{c}{} & \textbf{PSNR} $\uparrow$ & \textbf{SSIM} $\uparrow$ & \textbf{LPIPS} $\downarrow$ \\
\midrule
IBRNet$_{ft}$~\cite{wang2021ibrnet}     &    15.71           &    0.75           &        0.29         \\
MVSNeRF~\cite{mvsnerf}     &    18.37          &    \underline{0.81}           &        0.25       \\
VolSDF~\cite{volsdf}     &    14.18           &    0.62           &        0.35        \\
S-VolSDF~\cite{s-volsdf}\quad\quad\quad    &     19.67          &   0.71            &      0.30          \\
NeuSurf~\cite{neusurf} &  18.95 & 0.76  & 0.26 \\
SuGaR~\cite{sugar}      &    \textbf{21.32}           &    \textbf{0.83 }          &     \underline{0.24}           \\
\midrule
\rowcolor{lightgray}
\textbf{Ours}        &    \underline{20.78}           &    0.80          &     \textbf{0.20}          \\
\bottomrule
\bottomrule
\end{tabular}
\caption{\textbf{Quantitative NVS comparison on DTU}. Our method reaches state-of-the-art novel view synthesis quality by being on par with SuGaR~\cite{sugar}. Note that SuGaR achieves much lower surface quality (c.f. Tab.~\ref{tab:CD}) and shows artifacts in qualitative results (c.f. Fig.~\ref{fig:nvs}) in the given sparse-view setting.}
  \label{tab:nvs}
\end{table}

\begin{figure*}[!t]
    \centering
    \includegraphics[width=0.95\textwidth]{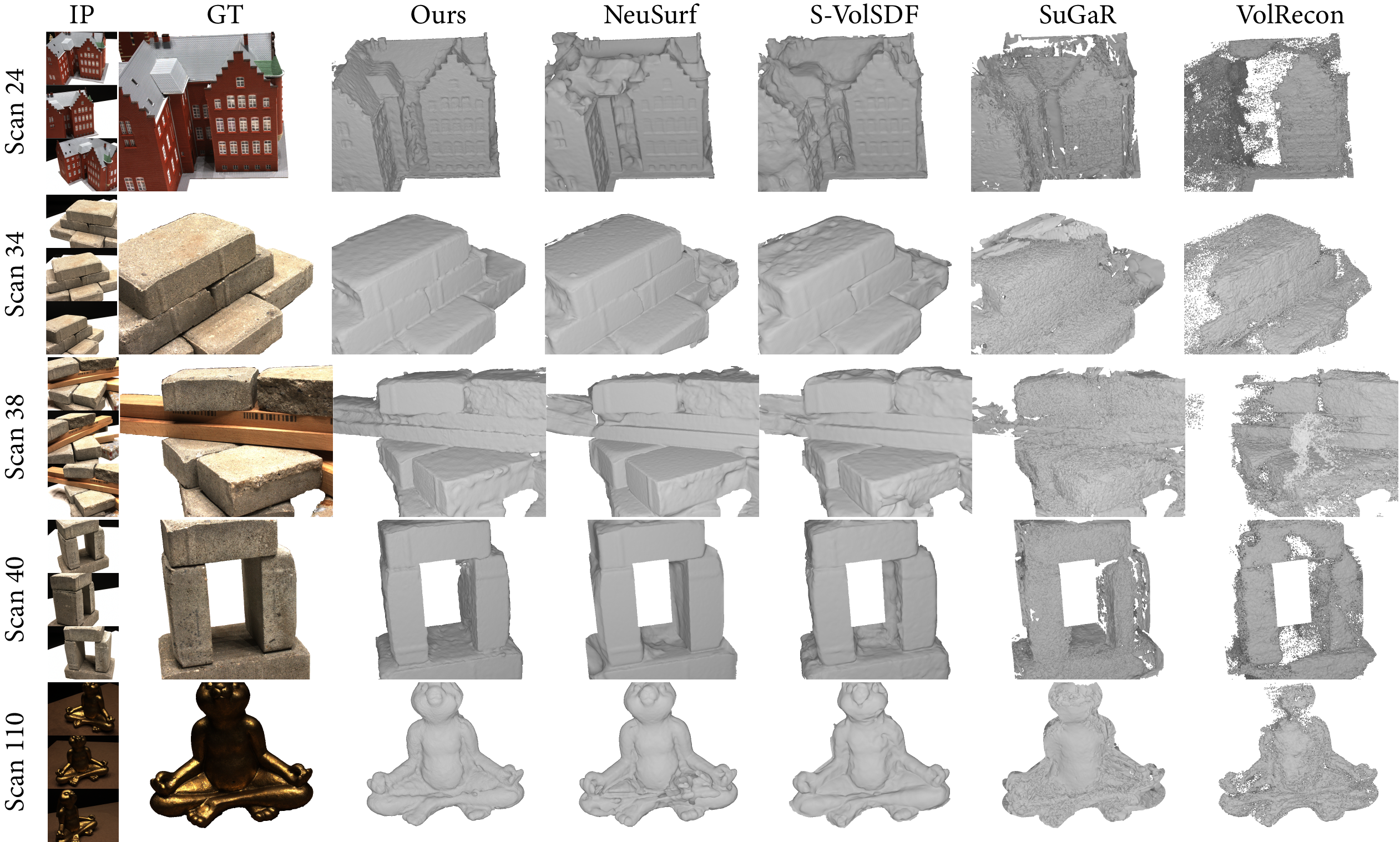}
    \caption{\textbf{Qualitative mesh reconstruction comparison on DTU}. Compared to previous state-of-the-art sparse-view methods, our reconstruction demonstrates superior completeness in regions with less view overlap. Our closest competitor is NeuSurf, which also reconstructs high quality surfaces on the object-centric DTU scenes. However, it fails to generalize to larger scenes (c.f. Fig.~\ref{fig:mesh-mip}).}
    \label{fig:sdf}
\end{figure*}

\vspace{-0.2cm}
\paragraph{Pseudo SDF Loss.}
We adapt a pseudo SDF loss $\mathcal{L}_{\mathrm{Pseu}}$ from previous work~\cite{geoneus} for our approach. It is minimized when the neural points $\mathbf{p}_i$ lie on the zero level set of the predicted SDF $\hat{s}$, i.e., on the surface.

\vspace{-0.1cm}
\paragraph{Optimization Objective.}
In total, we find the optimal parameters of $A_\mathrm{LP}$, $A_\mathrm{REG}$, $\mathbf{f}^g_i$, and $\mathbf{f}^a_i$ by minimizing:
\begin{equation}
    \mathcal{L}_{\mathrm{Rec}} = \mathcal{L}_{\mathrm{Ren}} + \lambda_{\mathrm{FC}} \cdot 
\mathcal{L}_{\mathrm{FC}} + \lambda_{\mathrm{Pseu}} \cdot \mathcal{L}_{\mathrm{Pseu}} + \lambda_{\mathrm{TV}} \cdot \mathcal{L}_{\mathrm{TV}}\textnormal{,}
\end{equation}
obtaining the final scene reconstruction.

\section{Experiments}
In this section, we compare Spurfies against previous methods for sparse-view surface reconstruction. We begin by introducing the used datasets, baselines and metrics, before presenting reconstruction results in Sec.~\ref{sec:reconstruction_results} and an ablation study with additional feature analysis in Sec.~\ref{sec:ablations}.

\paragraph{Datasets.}
In our evaluation process, we utilize the DTU dataset~\cite{dtu}, a comprehensive collection of object scans featuring 49 images per scan, each with a resolution of $576 \times 768$. The dataset provides point clouds, camera intrinsics, and poses, making it ideal for assessing sparse-view reconstruction methods. We observe that various approaches in the literature differ in their selection of input views for evaluation. Adopting the strategy employed by PixelNeRF~\cite{pixelnerf}, we take views 22, 25, and 28 from each scan as input, leaving 25 test views. The setup represents a dispersed view distribution with low overlap. For evaluation, we follow previous work S-VolSDF~\cite{s-volsdf}, using the same set of scans from the DTU dataset for our approach and all baseline methods.
We also evaluate on Mip-NeRF360~\cite{mipnerf}, a dataset of larger, unbounded scenes, captured from dense views. Since none of the previous works evaluates sparse-view reconstruction on Mip-NeRF360, we randomly choose a subset of three views to test. Note that none of our test scenes are in the training data of DUSt3R~\cite{dust3r}.

\begin{figure*}[t]
    \centering
    \includegraphics[width=0.95\textwidth]{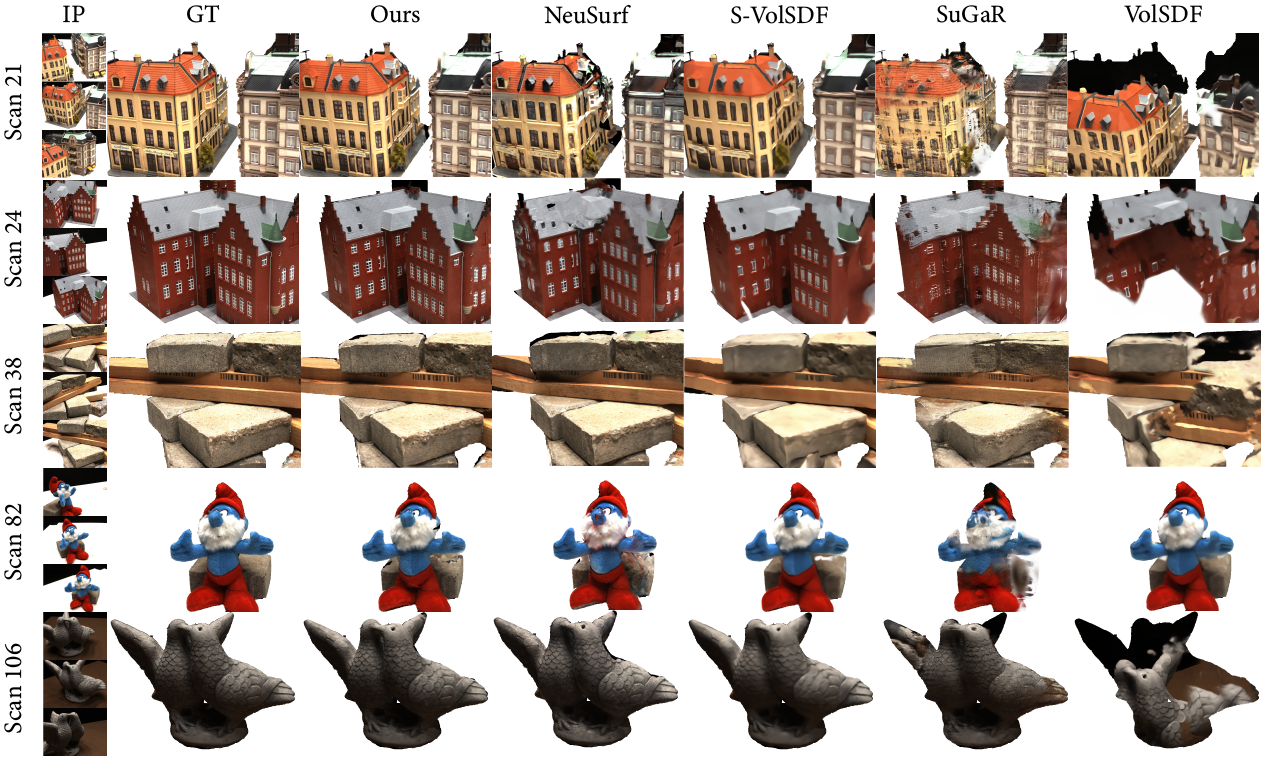}
    \caption{\textbf{Qualitative NVS comparison on DTU}. Our method demonstrates superior novel view synthesis quality relative to previous techniques. In comparison to NeuSurf~\cite{neusurf} and S-VolSDF~\cite{s-volsdf}, our method produces sharper details. The Gaussian reconstruction method SuGaR~\cite{sugar} and volume rendering approach VolSDF~\cite{volsdf} are not designed for sparse-view reconstruction and, therefore, show typical artifacts.}
    \label{fig:nvs}
\end{figure*}

\paragraph{Baselines.}
We compare our work against multiple state-of-the-art sparse-view reconstruction techniques: SparseNeuS~\cite{sparseneus}, VolRecon~\cite{volrecon}, NeuSurf~\cite{neusurf} and S-VolSDF~\cite{s-volsdf}. These methods leverage MVS priors predominantly trained on multi-view datasets such as DTU~\cite{dtu}, ensuring they operate within a similar data distribution. Additionally, we compare our approach with another point-based (dense-view) reconstruction method SuGaR~\cite{sugar} that employs Gaussians as the underlying 3D representation. To ensure a fair comparison, we initialize SuGaR~\cite{sugar} with point clouds obtained using DUSt3R~\cite{dust3r}, which improves its results. We also evaluate against surface reconstruction baselines directly from point clouds such as Poisson reconstruction~\cite{poisson}, Points2Surf~\cite{points2surf} and CAP-UDF~\cite{cap-udf}, directly applied to the DUSt3R~\cite{dust3r} point clouds. Further, we compare against dense volume rendering methods such as VolSDF~\cite{volsdf} and NeuS~\cite{neus}.

\paragraph{Metrics.}
We evaluate our technique against the baselines using both mesh reconstruction and novel view synthesis (NVS) metrics.
For mesh quality assessment, we employ the Chamfer Distance (CD), which measures the accuracy of the reconstructed mesh by calculating the average distance between points sampled from the mesh and the corresponding points in the ground-truth point cloud. For NVS, we utilize Peak Signal-to-Noise Ratio (PSNR), Structural Similarity Index Measure (SSIM), and Learned Perceptual Image Patch Similarity (LPIPS).

\paragraph{Implementation Details.}
For the neural point representation, we built upon codebases from previous work~\cite{simnp,pointnerf}.
Our training process for the local prior utilizes a batch size of 5. Each batch instance comprises 40,000 randomly sampled query points, equally distributed w.r.t. positive and negative signed distances, along with ~2,000 neural points. We train the geometry networks and the latent codes for 5,000 epochs. The total training time for the local prior is approximately 8 hours, utilizing a single A100 GPU. 

For sparse-view reconstruction, we optimize the parameters for 100,000 iterations using the Adam optimizer~\cite{adam} on a single A100 GPU, which takes ~3.5 hrs to converge depending upon the number of neural points. We refer to the supplementary material for more details.

\subsection{Reconstruction Results}

\label{sec:reconstruction_results}
\begin{figure}
    \centering
    \includegraphics[width=\linewidth]{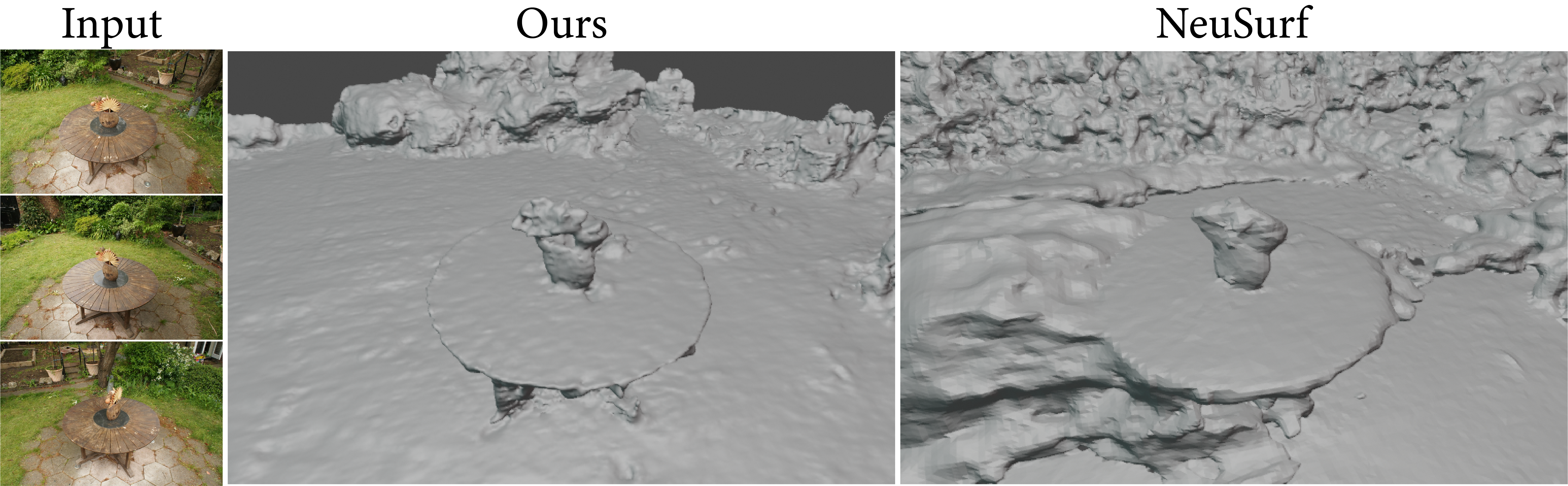}
    \caption{\textbf{Qualitative mesh reconstruction comparison on Mip-NeRF360~\cite{mipnerf}}. While NeuSurf~\cite{neusurf} also produces good results on object-centric scenes it fails on larger, unbounded scenes. In contrast, our local prior generalizes well to Mip-NeRF360. Please refer to the appendix for additional results.}
    \label{fig:mesh-mip}
\end{figure}

Our method demonstrates state-of-the-art quantitative results (Table~\ref{tab:CD}) in comparison to previous sparse-view and point-based reconstruction techniques. Qualitative comparisons are presented in Figure~\ref{fig:sdf} and the supplement.
Our reconstructions exhibit superior completeness and absence of floaters, resulting in enhanced reconstruction quality. Notably, while previous works utilize DTU for prior training and inference on the same data distribution, our prior is trained exclusively on synthetic data (Section~\ref{sec:lp}). Leveraging our local property, we evaluate our method on unbounded scenes from Mip-NeRF360. Figure~\ref{fig:mesh-mip} presents qualitative mesh comparison results with NeuSurf~\cite{neusurf}, illustrating that our method achieves significantly more accurate reconstructions. In the absence of ground-truth point clouds, our evaluation is limited to qualitative comparisons with more results in the supplementary material. 

Quantitative and qualitative results for novel view synthesis are presented in Table~\ref{tab:nvs} and Figure~\ref{fig:nvs}, respectively. Our method achieves state-of-the-art NVS results among volume rendering-based techniques. The results are comparable to the Gaussian splatting-based SuGaR~\cite{sugar}, despite SuGaR's inferior surface reconstruction quality.

\begin{figure}
    \centering
    \includegraphics[width=\linewidth]{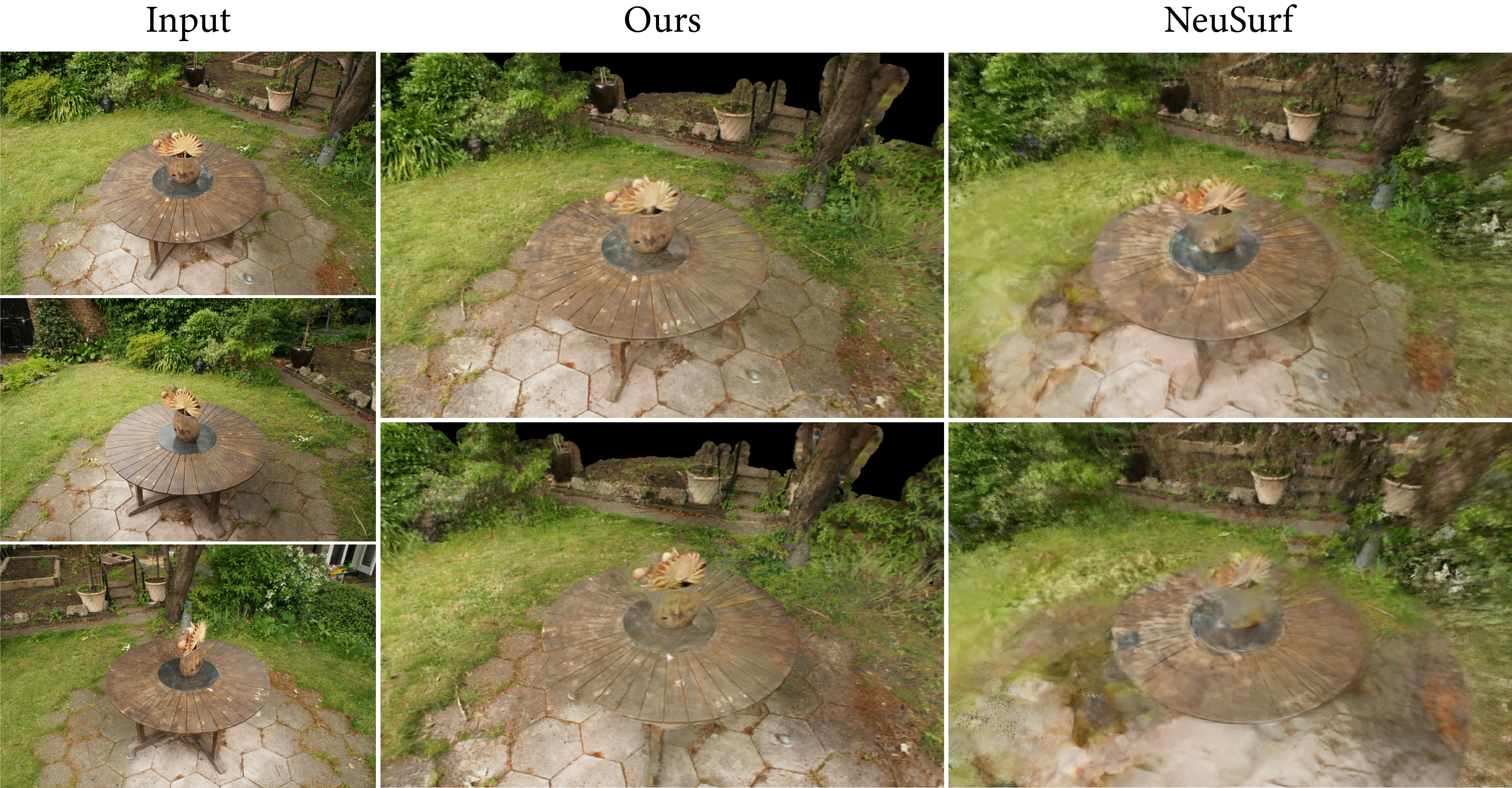}
    \caption{\textbf{Qualitative NVS comparison on Mip-NeRF360~\cite{mipnerf}}. Our geometry prior acts as effective regularization for appearance resulting in robust fitting of sparse views. Novel views from NeuSurf~\cite{neusurf} quickly deteriorate when moving away from training views, indicating overfitting.}
    \label{fig:enter-label}
\end{figure}

\begin{figure}[t]
    \centering
    \includegraphics[width=\linewidth, keepaspectratio]{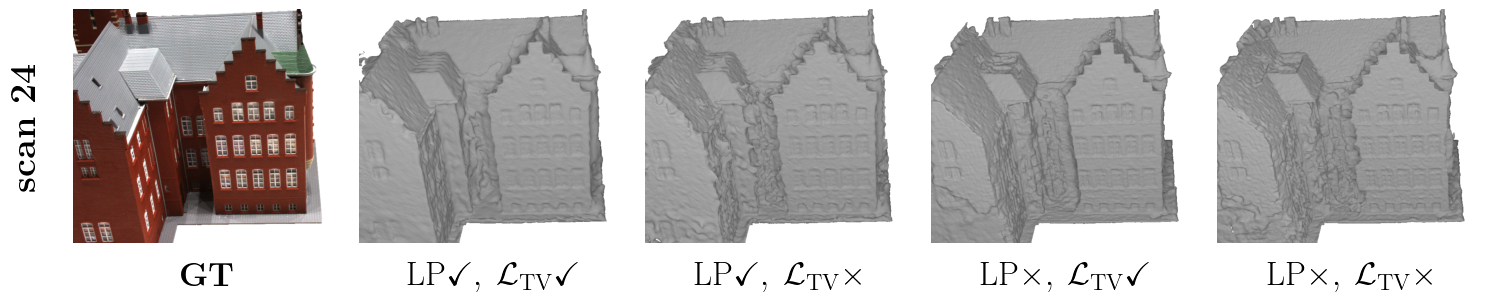}
    \caption{\textbf{Qualitative ablation} of our local prior (LP) and total variation $\mathcal{L}_{\text{TV}}$ regularization of geometry latent codes. It can be seen that both improve surface quality.}
    \label{fig:ablation}
\end{figure}

\begin{table}[t]
  \centering
\begin{tabular}{cc|c}
\toprule
\toprule
$\quad\text{Local Prior}$ \quad& \quad$\mathcal{L}_{\mathrm{TV}}$ \quad \quad & \quad \quad\textbf{Mean CD} \quad \quad$\downarrow$ \\
\midrule
$\times$ & $\times$ & 2.09 \\ 
$\times$ & $\checkmark$ & 1.91 \\ 
$\checkmark$ & $\times$ & 1.59 \\
$\checkmark$ & $\checkmark$ & \textbf{1.36} \\
\bottomrule
\bottomrule
\end{tabular}
\caption{\textbf{Quantitative ablation} of the proposed local prior and total variation regularization of geometry latent codes. Both contribute strongly to the reconstruction quality.}
  \label{ablation}
\end{table}

\begin{figure}[!t]
    \centering
    \includegraphics[width=\linewidth, keepaspectratio]{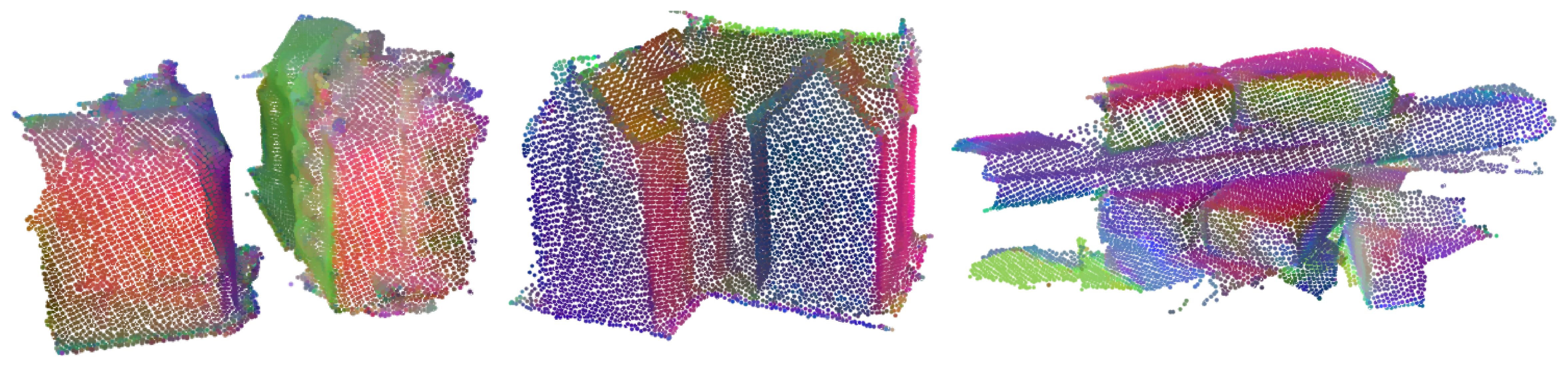}
    \caption{\textbf{Learned geometry latent codes} visualized via PCA. We observe similar features for points with same surface orientations.}
    \label{fig:feat-vis}
\end{figure}

\subsection{Ablation Study}
\label{sec:ablations}

We study the effect of our proposed local prior and the total variation regularization $\mathcal{L}_{\text{TV}}$ of the geometry latent codes with quantitative results in Table~\ref{ablation} and qualitative results in Figure~\ref{fig:ablation}. 
With the proposed point-based representation and regularizing losses from prior works (cf. Sec.~\ref{sec:svr}), Spurfies matches the performance of the state-of-the-art baseline NeuSurf~\cite{neusurf}. Leveraging the local prior obtained by training on synthetic data significantly enhances the level of detail in reconstructions. The total variation regularization on top mainly removes noise in under-constrained regions resulting in clean meshes.

\paragraph{Analysis.} In Figure~\ref{fig:feat-vis}, we visualize the learned geometry latent codes using Principal Component Analysis (PCA). It reveals a correlation between similar orientations and their corresponding geometry codes. By clustering the points based on their learnt latent codes we can also extract parts that are similarly oriented (c.f. supplementary material). 
\section{Conclusion}

We introduced Spurfies, a novel approach for sparse-view surface reconstruction that uses a distributed neural representation on point clouds and completely disentangles geometry and appearance. Our work shows that using a small subset of accurate synthetic ShapeNet data is sufficient to learn a high-quality local surface prior. 
We evaluated our method on the DTU and Mip-NeRF360 datasets, demonstrating state-of-the-art performance in mesh reconstruction from sparse views on DTU and that our method does generalize to large scenes. Our results indicate that Spurfies effectively mitigates the shape-radiance ambiguity inherent in sparse-view settings, producing high-quality surface reconstructions with sparse input views.
Our work provides an important insight for the future direction of the field, which is that the geometry prior is most important and can effectively be modeled by disentangling it from appearance and by training it on synthetic data.

{
    \small
    \bibliographystyle{ieeenat_fullname}
    \bibliography{main}

\begin{thebibliography}{55}
\providecommand{\natexlab}[1]{#1}
\providecommand{\url}[1]{\texttt{#1}}
\expandafter\ifx\csname urlstyle\endcsname\relax
  \providecommand{\doi}[1]{doi: #1}\else
  \providecommand{\doi}{doi: \begingroup \urlstyle{rm}\Url}\fi

\bibitem[Aan{\ae}s et~al.(2016)Aan{\ae}s, Jensen, Vogiatzis, Tola, and Dahl]{large}
Henrik Aan{\ae}s, Rasmus~Ramsb{\o}l Jensen, George Vogiatzis, Engin Tola, and Anders~Bjorholm Dahl.
\newblock Large-scale data for multiple-view stereopsis.
\newblock \emph{International Journal of Computer Vision}, 120:\penalty0 153--168, 2016.

\bibitem[Aliev et~al.(2020)Aliev, Sevastopolsky, Kolos, Ulyanov, and Lempitsky]{pbgraphics}
Kara-Ali Aliev, Artem Sevastopolsky, Maria Kolos, Dmitry Ulyanov, and Victor Lempitsky.
\newblock Neural point-based graphics.
\newblock In \emph{ECCV}, 2020.

\bibitem[Barron et~al.(2022)Barron, Mildenhall, Verbin, Srinivasan, and Hedman]{mipnerf}
Jonathan~T Barron, Ben Mildenhall, Dor Verbin, Pratul~P Srinivasan, and Peter Hedman.
\newblock Mip-nerf 360: Unbounded anti-aliased neural radiance fields.
\newblock In \emph{CVPR}, 2022.

\bibitem[Chabra et~al.(2020)]{deepls}
Rohan Chabra et~al.
\newblock Deep local shapes: Learning local sdf priors for detailed 3d reconstruction.
\newblock In \emph{ECCV}, 2020.

\bibitem[Chang et~al.(2015)Chang, Funkhouser, Guibas, Hanrahan, Huang, Li, Savarese, Savva, Song, Su, et~al.]{shapenet}
Angel~X Chang, Thomas Funkhouser, Leonidas Guibas, Pat Hanrahan, Qixing Huang, Zimo Li, Silvio Savarese, Manolis Savva, Shuran Song, Hao Su, et~al.
\newblock Shapenet: An information-rich 3d model repository.
\newblock \emph{arXiv}, 2015.

\bibitem[Charatan et~al.(2024)Charatan, Li, Tagliasacchi, and Sitzmann]{pixelsplat}
David Charatan, Sizhe Li, Andrea Tagliasacchi, and Vincent Sitzmann.
\newblock pixelsplat: 3d gaussian splats from image pairs for scalable generalizable 3d reconstruction.
\newblock In \emph{CVPR}, 2024.

\bibitem[Chen et~al.(2021)Chen, Xu, Zhao, Zhang, Xiang, Yu, and Su]{mvsnerf}
Anpei Chen, Zexiang Xu, Fuqiang Zhao, Xiaoshuai Zhang, Fanbo Xiang, Jingyi Yu, and Hao Su.
\newblock Mvsnerf: Fast generalizable radiance field reconstruction from multi-view stereo.
\newblock In \emph{CVPR}, 2021.

\bibitem[Chen et~al.(2024)Chen, Xu, Zheng, Zhuang, Pollefeys, Geiger, Cham, and Cai]{mvsplat}
Yuedong Chen, Haofei Xu, Chuanxia Zheng, Bohan Zhuang, Marc Pollefeys, Andreas Geiger, Tat-Jen Cham, and Jianfei Cai.
\newblock Mvsplat: Efficient 3d gaussian splatting from sparse multi-view images.
\newblock In \emph{ECCV}, 2024.

\bibitem[Das et~al.(2024)Das, Wewer, Yunus, Ilg, and Lenssen]{npg}
Devikalyan Das, Christopher Wewer, Raza Yunus, Eddy Ilg, and Jan~Eric Lenssen.
\newblock Neural parametric gaussians for monocular non-rigid object reconstruction.
\newblock In \emph{CVPR}, 2024.

\bibitem[Erler et~al.(2020)Erler, Guerrero, Ohrhallinger, Mitra, and Wimmer]{points2surf}
Philipp Erler, Paul Guerrero, Stefan Ohrhallinger, Niloy~J Mitra, and Michael Wimmer.
\newblock Points2surf learning implicit surfaces from point clouds.
\newblock In \emph{ECCV}, 2020.

\bibitem[Fu et~al.(2022)Fu, Xu, Ong, and Tao]{geoneus}
Qiancheng Fu, Qingshan Xu, Yew-Soon Ong, and Wenbing Tao.
\newblock Geo-neus: Geometry-consistent neural implicit surfaces learning for multi-view reconstruction.
\newblock In \emph{NeurIPS}, 2022.

\bibitem[Gropp et~al.(2020)Gropp, Yariv, Haim, Atzmon, and Lipman]{eik}
Amos Gropp, Lior Yariv, Niv Haim, Matan Atzmon, and Yaron Lipman.
\newblock Implicit geometric regularization for learning shapes.
\newblock In \emph{Machine Learning and Systems}, 2020.

\bibitem[Gu et~al.(2020)Gu, Fan, Zhu, Dai, Tan, and Tan]{casmvsnet}
Xiaodong Gu, Zhiwen Fan, Siyu Zhu, Zuozhuo Dai, Feitong Tan, and Ping Tan.
\newblock Cascade cost volume for high-resolution multi-view stereo and stereo matching.
\newblock In \emph{CVPR}, 2020.

\bibitem[Gu{\'e}don and Lepetit(2024)]{sugar}
Antoine Gu{\'e}don and Vincent Lepetit.
\newblock Sugar: Surface-aligned gaussian splatting for efficient 3d mesh reconstruction and high-quality mesh rendering.
\newblock In \emph{CVPR}, 2024.

\bibitem[Huang et~al.(2024{\natexlab{a}})Huang, Yu, Chen, Geiger, and Gao]{2dgs}
Binbin Huang, Zehao Yu, Anpei Chen, Andreas Geiger, and Shenghua Gao.
\newblock 2d gaussian splatting for geometrically accurate radiance fields.
\newblock In \emph{SIGGRAPH}, 2024{\natexlab{a}}.

\bibitem[Huang et~al.(2024{\natexlab{b}})Huang, Wu, Zhou, Gao, Gu, and Liu]{neusurf}
Han Huang, Yulun Wu, Junsheng Zhou, Ge Gao, Ming Gu, and Yu-Shen Liu.
\newblock Neusurf: On-surface priors for neural surface reconstruction from sparse input views.
\newblock In \emph{AAAI}, 2024{\natexlab{b}}.

\bibitem[Jensen et~al.(2014)Jensen, Dahl, Vogiatzis, Tola, and Aan{\ae}s]{dtu}
Rasmus Jensen, Anders Dahl, George Vogiatzis, Engil Tola, and Henrik Aan{\ae}s.
\newblock Large scale multi-view stereopsis evaluation.
\newblock In \emph{CVPR}, 2014.

\bibitem[Kazhdan et~al.(2006)Kazhdan, Bolitho, and Hoppe]{poisson}
Michael Kazhdan, Matthew Bolitho, and Hugues Hoppe.
\newblock Poisson surface reconstruction.
\newblock In \emph{Proceedings of the fourth Eurographics symposium on Geometry processing}, 2006.

\bibitem[Kerbl et~al.(2023)Kerbl, Kopanas, Leimk{\"u}hler, and Drettakis]{3dgs}
Bernhard Kerbl, Georgios Kopanas, Thomas Leimk{\"u}hler, and George Drettakis.
\newblock 3d gaussian splatting for real-time radiance field rendering.
\newblock In \emph{ACM TOG}, 2023.

\bibitem[Kingma and Ba(2014)]{adam}
Diederik~P Kingma and Jimmy Ba.
\newblock Adam: A method for stochastic optimization.
\newblock \emph{arXiv}, 2014.

\bibitem[Li et~al.(2023)Li, M\"uller, Evans, Taylor, Unberath, Liu, and Lin]{neuralangelo}
Zhaoshuo Li, Thomas M\"uller, Alex Evans, Russell~H Taylor, Mathias Unberath, Ming-Yu Liu, and Chen-Hsuan Lin.
\newblock Neuralangelo: High-fidelity neural surface reconstruction.
\newblock In \emph{CVPR}, 2023.

\bibitem[{Liang} et~al.(2022){Liang}, {Zhang}, {Li}, {Yang}, {Guan}, and {Vijaykumar}]{spidr}
Ruofan {Liang}, Jiahao {Zhang}, Haoda {Li}, Chen {Yang}, Yushi {Guan}, and Nandita {Vijaykumar}.
\newblock {SPIDR: SDF-based Neural Point Fields for Illumination and Deformation}.
\newblock In \emph{arXiv}, 2022.

\bibitem[Liang et~al.(2023)Liang, He, and Chen]{retr}
Yixun Liang, Hao He, and Yingcong Chen.
\newblock Retr: Modeling rendering via transformer for generalizable neural surface reconstruction.
\newblock In \emph{NeurIPS}, 2023.

\bibitem[Long et~al.(2022)Long, Lin, Wang, Komura, and Wang]{sparseneus}
Xiaoxiao Long, Cheng Lin, Peng Wang, Taku Komura, and Wenping Wang.
\newblock Sparseneus: Fast generalizable neural surface reconstruction from sparse views.
\newblock In \emph{ECCV}, 2022.

\bibitem[Luiten et~al.(2024)Luiten, Kopanas, Leibe, and Ramanan]{dynamicgaussians}
Jonathon Luiten, Georgios Kopanas, Bastian Leibe, and Deva Ramanan.
\newblock Dynamic 3d gaussians: Tracking by persistent dynamic view synthesis.
\newblock In \emph{3DV}, 2024.

\bibitem[Mildenhall et~al.(2020)Mildenhall, Srinivasan, Tancik, Barron, Ramamoorthi, and Ng]{nerf}
Ben Mildenhall, Pratul~P. Srinivasan, Matthew Tancik, Jonathan~T. Barron, Ravi Ramamoorthi, and Ren Ng.
\newblock {NeRF}: Representing scenes as neural radiance fields for view synthesis.
\newblock In \emph{ECCV}, 2020.

\bibitem[M\"uller et~al.(2022)M\"uller, Evans, Schied, and Keller]{instantngp}
Thomas M\"uller, Alex Evans, Christoph Schied, and Alexander Keller.
\newblock Instant neural graphics primitives with a multiresolution hash encoding.
\newblock In \emph{ACM TOG}, 2022.

\bibitem[Niemeyer et~al.(2022)Niemeyer, Barron, Mildenhall, Sajjadi, Geiger, and Radwan]{regnerf}
Michael Niemeyer, Jonathan~T Barron, Ben Mildenhall, Mehdi~SM Sajjadi, Andreas Geiger, and Noha Radwan.
\newblock Regnerf: Regularizing neural radiance fields for view synthesis from sparse inputs.
\newblock In \emph{Proceedings of the IEEE/CVF Conference on Computer Vision and Pattern Recognition}, pages 5480--5490, 2022.

\bibitem[Park et~al.(2019)Park, Florence, Straub, Newcombe, and Lovegrove]{deepsdf}
Jeong~Joon Park, Peter Florence, Julian Straub, Richard Newcombe, and Steven Lovegrove.
\newblock Deepsdf: Learning continuous signed distance functions for shape representation.
\newblock In \emph{CVPR}, 2019.

\bibitem[Rakhimov et~al.(2022)Rakhimov, Ardelean, Lempitsky, and Burnaev]{pbgraphics+}
Ruslan Rakhimov, Andrei-Timotei Ardelean, Victor Lempitsky, and Evgeny Burnaev.
\newblock {NPBG++}: Accelerating neural point-based graphics.
\newblock In \emph{CVPR}, 2022.

\bibitem[Ren et~al.(2023)Ren, Zhang, Pollefeys, S{\"u}sstrunk, and Wang]{volrecon}
Yufan Ren, Tong Zhang, Marc Pollefeys, Sabine S{\"u}sstrunk, and Fangjinhua Wang.
\newblock Volrecon: Volume rendering of signed ray distance functions for generalizable multi-view reconstruction.
\newblock In \emph{CVPR}, 2023.

\bibitem[Schröppel et~al.(2024)Schröppel, Wewer, Lenssen, Ilg, and Brox]{npcd}
Philipp Schröppel, Christopher Wewer, Jan~Eric Lenssen, Eddy Ilg, and Thomas Brox.
\newblock Neural point cloud diffusion for disentangled 3d shape and appearance generation.
\newblock In \emph{CVPR}, 2024.

\bibitem[Turkulainen et~al.(2024)Turkulainen, Ren, Melekhov, Seiskari, Rahtu, and Kannala]{dnsplatter}
Matias Turkulainen, Xuqian Ren, Iaroslav Melekhov, Otto Seiskari, Esa Rahtu, and Juho Kannala.
\newblock Dn-splatter: Depth and normal priors for gaussian splatting and meshing.
\newblock In \emph{arXiv}, 2024.

\bibitem[Wang et~al.(2022)Wang, Wang, Long, Theobalt, Komura, Liu, and Wang]{neuris}
Jiepeng Wang, Peng Wang, Xiaoxiao Long, Christian Theobalt, Taku Komura, Lingjie Liu, and Wenping Wang.
\newblock Neuris: Neural reconstruction of indoor scenes using normal priors.
\newblock In \emph{ECCV}, 2022.

\bibitem[Wang et~al.(2021{\natexlab{a}})Wang, Liu, Liu, Theobalt, Komura, and Wang]{neus}
Peng Wang, Lingjie Liu, Yuan Liu, Christian Theobalt, Taku Komura, and Wenping Wang.
\newblock Neus: Learning neural implicit surfaces by volume rendering for multi-view reconstruction.
\newblock In \emph{NeurIPS}, 2021{\natexlab{a}}.

\bibitem[Wang et~al.(2021{\natexlab{b}})Wang, Wang, Genova, Srinivasan, Zhou, Barron, Martin-Brualla, Snavely, and Funkhouser]{wang2021ibrnet}
Qianqian Wang, Zhicheng Wang, Kyle Genova, Pratul Srinivasan, Howard Zhou, Jonathan~T. Barron, Ricardo Martin-Brualla, Noah Snavely, and Thomas Funkhouser.
\newblock Ibrnet: Learning multi-view image-based rendering.
\newblock In \emph{CVPR}, 2021{\natexlab{b}}.

\bibitem[Wang et~al.(2023{\natexlab{a}})Wang, Leroy, Cabon, Chidlovskii, and Revaud]{dust3r}
Shuzhe Wang, Vincent Leroy, Yohann Cabon, Boris Chidlovskii, and Jerome Revaud.
\newblock Dust3r: Geometric 3d vision made easy.
\newblock In \emph{arXiv}, 2023{\natexlab{a}}.

\bibitem[Wang et~al.(2023{\natexlab{b}})Wang, Han, Habermann, Daniilidis, Theobalt, and Liu]{neus2}
Yiming Wang, Qin Han, Marc Habermann, Kostas Daniilidis, Christian Theobalt, and Lingjie Liu.
\newblock Neus2: Fast learning of neural implicit surfaces for multi-view reconstruction.
\newblock In \emph{ICCV}, 2023{\natexlab{b}}.

\bibitem[Wewer et~al.(2023)Wewer, Ilg, Schiele, and Lenssen]{simnp}
Christopher Wewer, Eddy Ilg, Bernt Schiele, and Jan~Eric Lenssen.
\newblock {SimNP}: Learning self-similarity priors between neural points.
\newblock In \emph{ICCV}, 2023.

\bibitem[Wewer et~al.(2024)Wewer, Raj, Ilg, Schiele, and Lenssen]{latentsplat}
Christopher Wewer, Kevin Raj, Eddy Ilg, Bernt Schiele, and Jan~Eric Lenssen.
\newblock latentsplat: Autoencoding variational gaussians for fast generalizable 3d reconstruction.
\newblock In \emph{ECCV}, 2024.

\bibitem[Wu et~al.(2024{\natexlab{a}})Wu, Yi, Fang, Xie, Zhang, Wei, Liu, Tian, and Wang]{4dgs}
Guanjun Wu, Taoran Yi, Jiemin Fang, Lingxi Xie, Xiaopeng Zhang, Wei Wei, Wenyu Liu, Qi Tian, and Xinggang Wang.
\newblock 4d gaussian splatting for real-time dynamic scene rendering.
\newblock In \emph{CVPR}, 2024{\natexlab{a}}.

\bibitem[Wu et~al.(2023{\natexlab{a}})Wu, Graikos, and Samaras]{s-volsdf}
Haoyu Wu, Alexandros Graikos, and Dimitris Samaras.
\newblock S-volsdf: Sparse multi-view stereo regularization of neural implicit surfaces.
\newblock In \emph{ICCV}, 2023{\natexlab{a}}.

\bibitem[Wu et~al.(2024{\natexlab{b}})Wu, Mildenhall, Henzler, Park, Gao, Watson, Srinivasan, Verbin, Barron, Poole, and Holynski]{Wu2024CVPR}
Rundi Wu, Ben Mildenhall, Philipp Henzler, Keunhong Park, Ruiqi Gao, Daniel Watson, Pratul~P. Srinivasan, Dor Verbin, Jonathan~T. Barron, Ben Poole, and Aleksander Holynski.
\newblock Reconfusion: 3d reconstruction with diffusion priors.
\newblock In \emph{CVPR}, 2024{\natexlab{b}}.

\bibitem[Wu et~al.(2023{\natexlab{b}})Wu, Wang, Pan, Xu, Theobalt, Liu, and Lin]{voxurf}
Tong Wu, Jiaqi Wang, Xingang Pan, Xudong Xu, Christian Theobalt, Ziwei Liu, and Dahua Lin.
\newblock Voxurf: Voxel-based efficient and accurate neural surface reconstruction.
\newblock In \emph{ICLR}, 2023{\natexlab{b}}.

\bibitem[Xu et~al.(2023)Xu, Guan, Wang, Liu, Zeng, Wang, and Yang]{c2f2neus}
Luoyuan Xu, Tao Guan, Yuesong Wang, Wenkai Liu, Zhaojie Zeng, Junle Wang, and Wei Yang.
\newblock C2f2neus: Cascade cost frustum fusion for high fidelity and generalizable neural surface reconstruction.
\newblock In \emph{ICCV}, 2023.

\bibitem[Xu et~al.(2022)Xu, Xu, Philip, Bi, Shu, Sunkavalli, and Neumann]{pointnerf}
Qiangeng Xu, Zexiang Xu, Julien Philip, Sai Bi, Zhixin Shu, Kalyan Sunkavalli, and Ulrich Neumann.
\newblock {Point-NeRF}: Point-based neural radiance fields.
\newblock In \emph{CVPR}, 2022.

\bibitem[Yariv et~al.(2021)Yariv, Gu, Kasten, and Lipman]{volsdf}
Lior Yariv, Jiatao Gu, Yoni Kasten, and Yaron Lipman.
\newblock Volume rendering of neural implicit surfaces.
\newblock In \emph{NeurIPS}, 2021.

\bibitem[Yariv et~al.(2023)Yariv, Hedman, Reiser, Verbin, Srinivasan, Szeliski, Barron, and Mildenhall]{bakedsdf}
Lior Yariv, Peter Hedman, Christian Reiser, Dor Verbin, Pratul~P. Srinivasan, Richard Szeliski, Jonathan~T. Barron, and Ben Mildenhall.
\newblock Bakedsdf: Meshing neural sdfs for real-time view synthesis.
\newblock In \emph{ACM TOG}, 2023.

\bibitem[Yu et~al.(2021)Yu, Ye, Tancik, and Kanazawa]{pixelnerf}
Alex Yu, Vickie Ye, Matthew Tancik, and Angjoo Kanazawa.
\newblock pixelnerf: Neural radiance fields from one or few images.
\newblock In \emph{CVPR}, 2021.

\bibitem[Yu et~al.(2022)Yu, Peng, Niemeyer, Sattler, and Geiger]{monosdf}
Zehao Yu, Songyou Peng, Michael Niemeyer, Torsten Sattler, and Andreas Geiger.
\newblock Monosdf: Exploring monocular geometric cues for neural implicit surface reconstruction.
\newblock In \emph{NeurIPS}, 2022.

\bibitem[Zhang et~al.(2022)Zhang, Yao, Li, Fang, McKinnon, Tsin, and Quan]{regsdf}
Jingyang Zhang, Yao Yao, Shiwei Li, Tian Fang, David McKinnon, Yanghai Tsin, and Long Quan.
\newblock Critical regularizations for neural surface reconstruction in the wild.
\newblock In \emph{CVPR}, 2022.

\bibitem[Zhang et~al.(2023)Zhang, Li, Luo, Fang, and Yao]{vismvsnet}
Jingyang Zhang, Shiwei Li, Zixin Luo, Tian Fang, and Yao Yao.
\newblock Vis-mvsnet: Visibility-aware multi-view stereo network.
\newblock In \emph{ICCV}, 2023.

\bibitem[Zhang et~al.(2020)Zhang, Riegler, Snavely, and Koltun]{nerf++}
Kai Zhang, Gernot Riegler, Noah Snavely, and Vladlen Koltun.
\newblock Nerf++: Analyzing and improving neural radiance fields.
\newblock In \emph{arXiv}, 2020.

\bibitem[Zhao et~al.(2022)Zhao, Jiang, Yao, Zhang, Wang, Dai, Zhong, Zhang, Wu, Xu, and Yu]{instantnsr}
Fuqiang Zhao, Yuheng Jiang, Kaixin Yao, Jiakai Zhang, Liao Wang, Haizhao Dai, Yuhui Zhong, Yingliang Zhang, Minye Wu, Lan Xu, and Jingyi Yu.
\newblock Human performance modeling and rendering via neural animated mesh.
\newblock In \emph{ACM TOG}, 2022.

\bibitem[Zhou et~al.(2024)Zhou, Ma, Li, Liu, Fang, and Han]{cap-udf}
Junsheng Zhou, Baorui Ma, Shujuan Li, Yu-Shen Liu, Yi Fang, and Zhizhong Han.
\newblock Cap-udf: Learning unsigned distance functions progressively from raw point clouds with consistency-aware field optimization.
\newblock In \emph{IEEE TPAMI}, 2024.

\end{thebibliography}
}

\setcounter{section}{0}
\renewcommand\thesection{\Alph{section}}
\setcounter{figure}{0}
\setcounter{page}{1}
\maketitlesupplementary



\section{Architectures}

Our method employs four Multi-Layer Perceptrons (MLPs) and two sets of learnable latent codes. The MLPs are: $\mathit{A}_{\mathrm{LP}}$, $\mathit{G}_{\mathrm{LP}}$, used for local processing, signed distance regressor $\mathit{G}_{\mathrm{REG}}$, and Radiance regressor $\mathit{A}_{\mathrm{REG}}$. The latent codes are for color $f^a$ and geometry $f^g$. Here are the details of these components:

\paragraph{Latent codes:} The color latent codes $f^a \in \mathbb{R}^{64}$ and the geometry latent codes $f^g \in \mathbb{R}^{32}$ are both initialized from a normal distribution with variance $1\mathrm{e}{-4}$.

\paragraph{Radiance Local Processing $\mathit{A}_{\mathrm{LP}}$:} This MLP comprises four linear layers with intermediate dimensions of 128. It takes as input the color latent codes $f^a$ and the relative distance of query points to the neural points, with positional encoding using 6 frequencies.

\paragraph{Raidance Regression $\mathit{A}_{\mathrm{REG}}$:} processes the aggregated color latent codes along with the view direction (without positional encoding). It consists of three linear layers with dimensions [259, 128, 3].

\paragraph{Geometry Local Processing $\mathit{G}_{\mathrm{LP}}$:} This MLP also has four linear layers with intermediate dimensions of 128. It processes the geometry latent codes $f^g$ and the relative distance of query points to the neural points, without any positional encoding. This MLP is frozen after learning the local geometry prior and remains unchanged during sparse view surface reconstruction.

\paragraph{Signed Distance Regression $\mathit{G}_{\mathrm{REG}}$:} This consists of a single frozen linear layer that maps the processed geometry latent codes to an SDF value.

\begin{figure}[!t]
    \centering
    \includegraphics[width=0.5\textwidth]{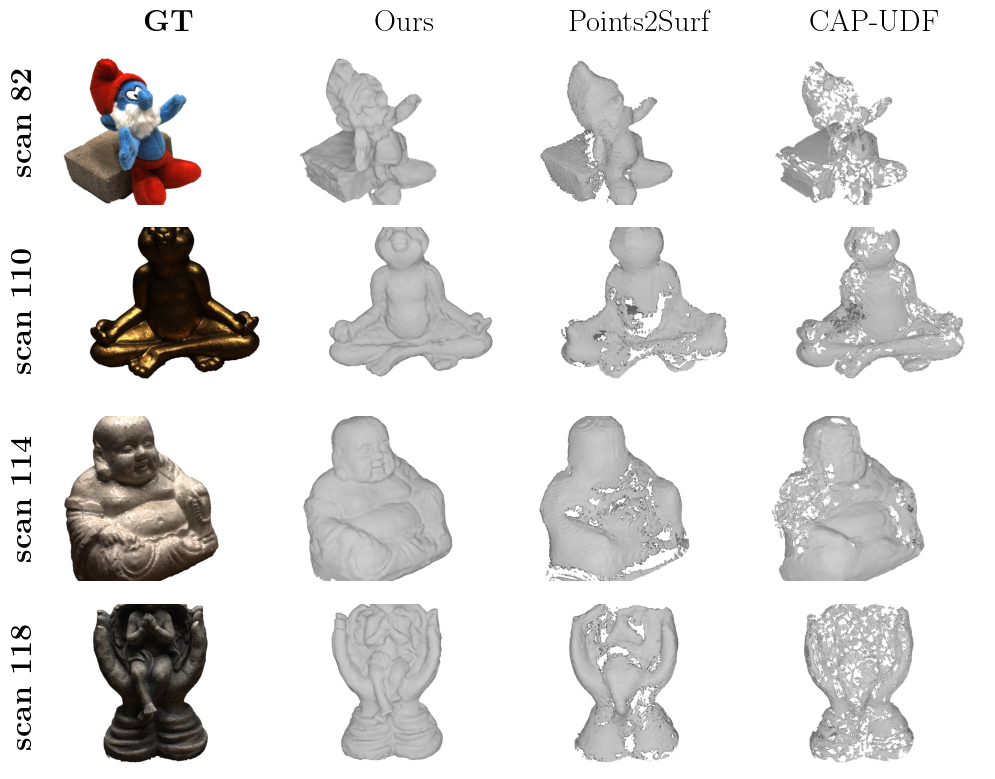}
    \caption{\textbf{Qualitative comparison} of mesh reconstruction with the point-based mesh reconstruction methods. In contrast to our approach, point-based mesh reconstruction methods often show missing areas, even when initialized with DUST3R~\cite{dust3r} point clouds.}
    \label{fig:pts_recon}  
\end{figure}

\begin{figure}[!t]
    \centering
    \includegraphics[width=0.4\textwidth]{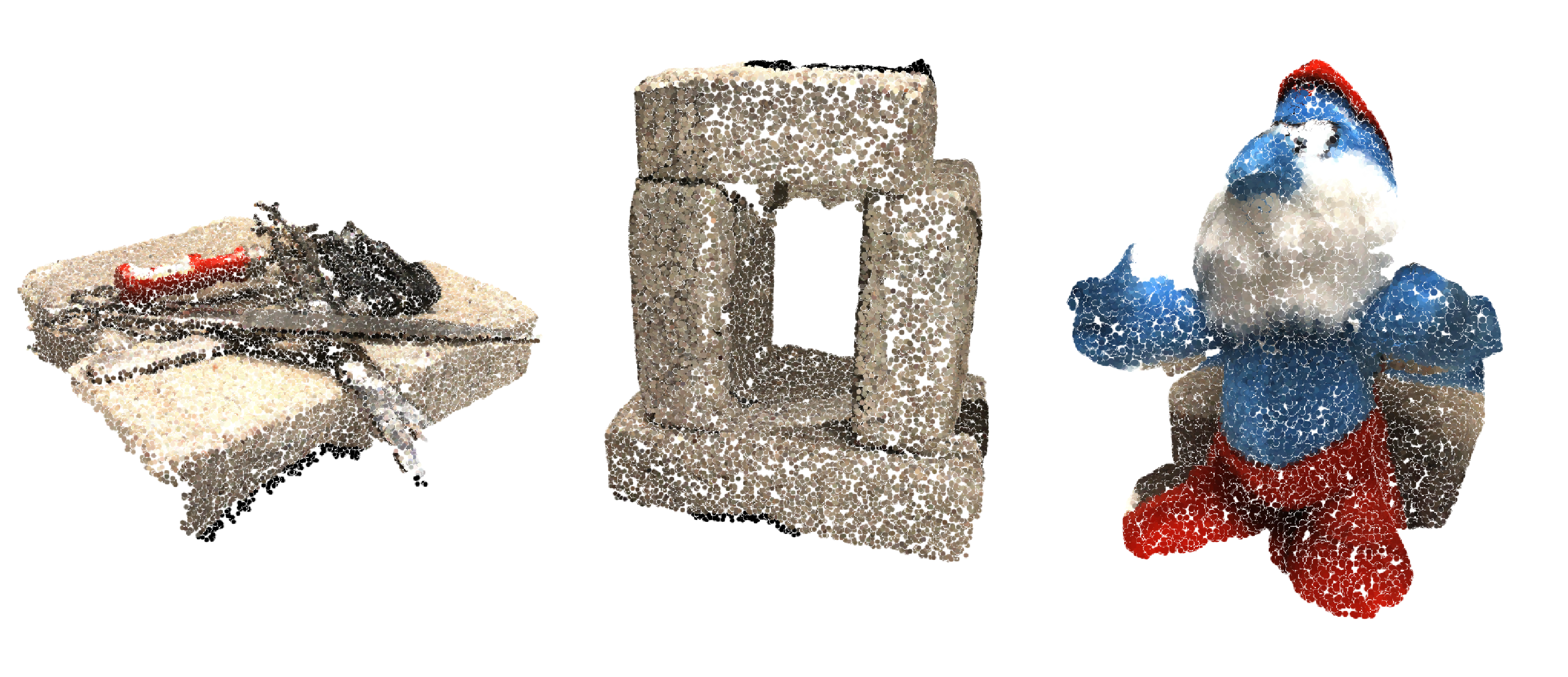}
    \caption{Sampled points from the reconstructed mesh on few scans from DTU dataset.}
    \label{fig:pc-vis}
\end{figure}

\section{Loss Functions}
\paragraph{Feature Consistency Loss $\mathcal{L}_\mathrm{FC}$~\cite{neusurf}:}
First, we estimate a set of surface points $\hat{\mathcal{P}} = \{\mathbf{p} | \mathbf{p} = \mathbf{x}_{i,u,v}(t^*)\}$ by finding zero-crossings $t^*$  along each ray $\mathbf{x}_{i,u,v}(t)$ that are computed using linear interpolation between adjacent samples:
\begin{equation}
   t^* = \frac{\hat{s}(\mathbf{x}(t_j)) t_{j+1} - \hat{s}(\mathbf{x}(t_{j+1})) t_j}{\hat{s}(\mathbf{r}(t_j)) - \hat{s}(\mathbf{x}(t_{j+1}))} \mathrm{\,.}
\end{equation}
where $t_j$ is estimated using $\hat{s}(\vx (t_j)) \cdot \hat{s}(\vx (t_{j+1})) < 0$.
We then define the photo-consistency loss as:
\begin{equation}
   \mathcal{L}_{\mathrm{FC}} = \frac{1}{|\hat{\mathcal{P}}||\mathcal{I}|} \sum_{p_i \in \hat{\mathcal{P}}} \sum_{\pi_j \in \Pi{}} \|f_{\phi}(\pi_j(p_i)) - f_{\phi}(\pi_0(p_i))\|_1,
\end{equation}
where $\Pi$ is the set of the projection matrices for the images $\mathcal{I}$, with $\mathcal{I}_0$ being the reference view, and $f_{\phi}$ computed with VisMVSNet~\cite{vismvsnet}. 

\paragraph{Pseudo loss $\mathcal{L}_\mathrm{pseu}$}:
We estimate surface points using rendering weights~\cite{geoneus}. This approach ensures that the estimated points have a SDF value close to zero, effectively lying on the surface.
We compute the estimated surface point location $t^*$ along a ray $\mathbf{x}(t)$ as a weighted average of sample positions:
\begin{equation}
t^* = \sum_i \frac{w_i \cdot t_i}{ \sum w_i },
\end{equation}
where $w_i$ are the rendering weights and $t_i$ are the sample depths along the ray.
Using these estimated surface points, we introduce the pseudo ground-truth loss:
\begin{equation}
\mathcal{L}_{\mathrm{Pseu}} = \frac{1}{N} \sum || \hat{s}(\mathbf{x}(t^*)) || \mathrm{\,.}
\end{equation}

\section{Implementation Details}
\paragraph{Training:} For sparse view reconstruction, we optimize a composite loss function:

\begin{equation}
L_{total} = \mathcal{L}_{\mathrm{ren}} + \lambda_{\mathrm{fc}} \cdot 
\mathcal{L}_{\mathrm{fc}} + \lambda_{\mathrm{pseu}} \cdot \mathcal{L}_{\mathrm{pseu}} + \lambda_{\mathrm{TV}} \cdot \mathcal{L}_{\mathrm{TV}} \textnormal{,}
\end{equation}
where $\lambda_{\mathrm{fc}} = 0.5$, $\lambda_{\mathrm{pseu}} = 0.5$, and $\lambda_{\mathrm{TV}} = 0.01$. We train the model for 100,000 iterations using the Adam optimizer~\cite{adam} on a single A100 GPU.

We implemented efficient querying of K neural neighbors implemented using a GPU-accelerated VoxelGrid approach~\cite{simnp, pointnerf}. The VoxelGrid parameters are configured as follows:

\begin{figure*}[!t]
    \centering
    \includegraphics[width=\textwidth]{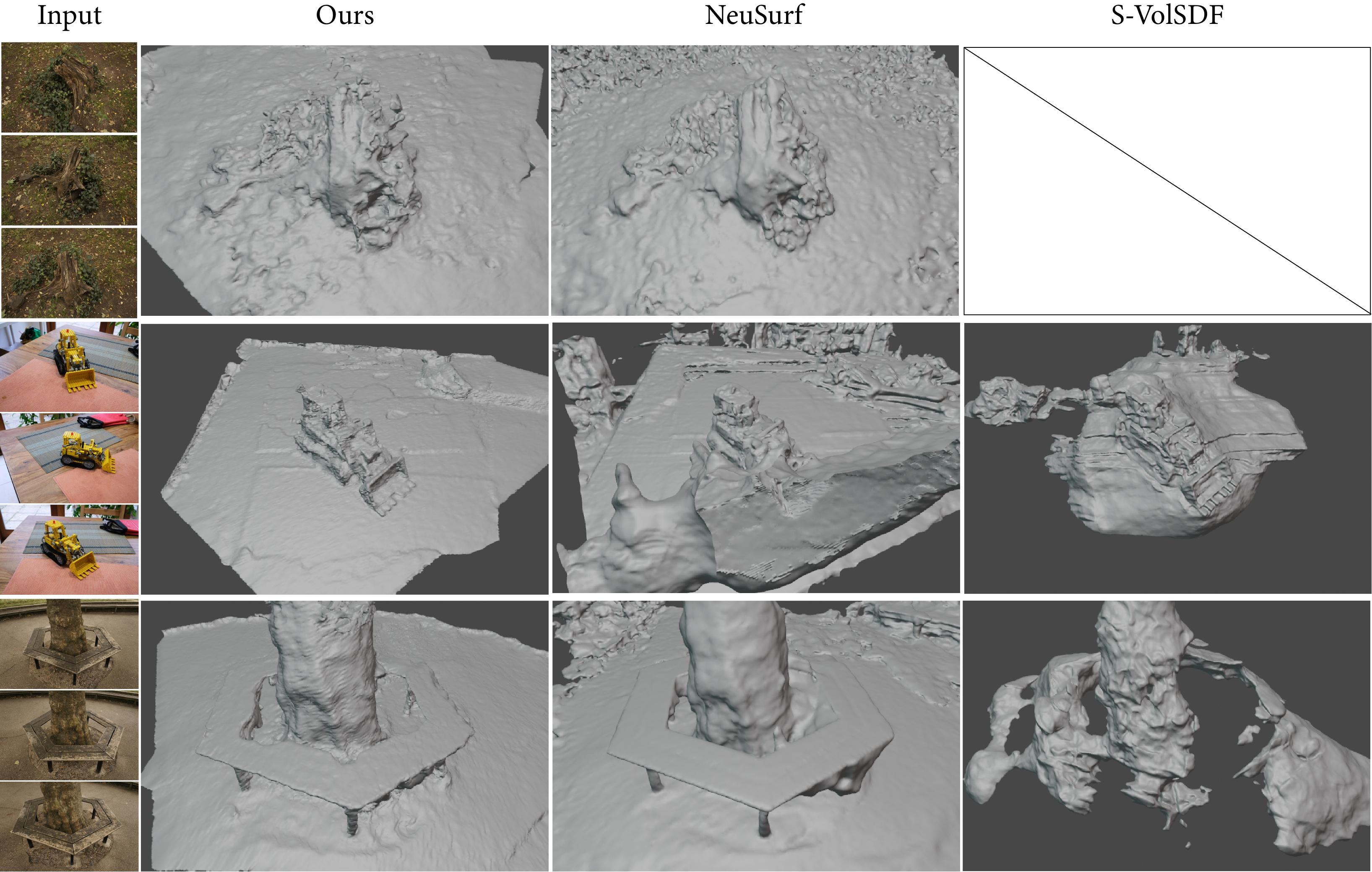}
    \caption{\textbf{Qualitative mesh reconstruction on Mip-NeRF 360~\cite{mipnerf}}. Compared to previous sparse view methods, we can achieve much better reconstruction on larger, unbounded scenes. S-VolSDF completely failed on the stump scene.}
    \label{fig:mip-extra}
\end{figure*}

\begin{figure*}[!t]
    \centering
    \includegraphics[width=\textwidth]{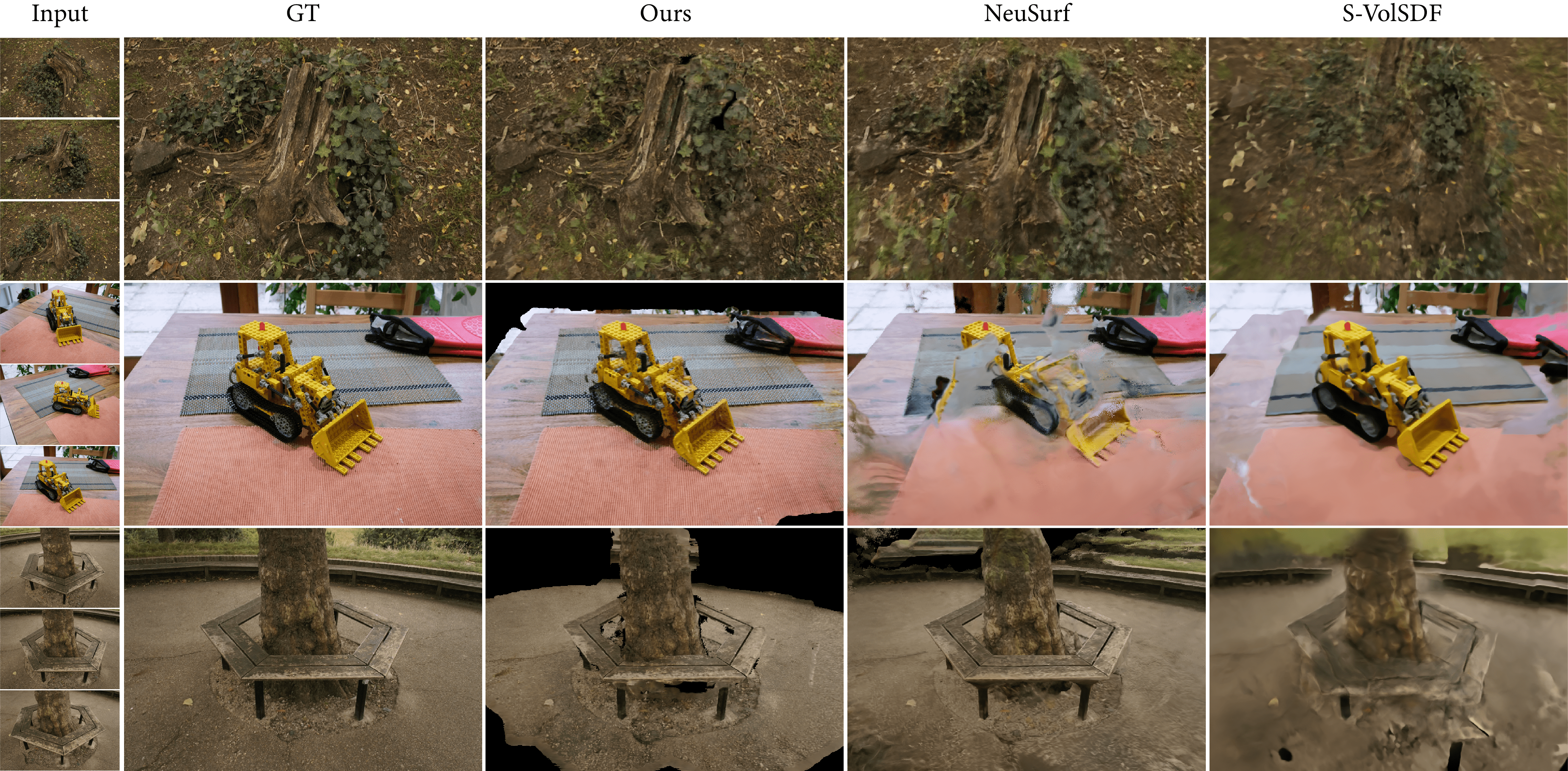}
    \caption{\textbf{Qualitative NVS on Mip-NeRF 360~\cite{mipnerf}}. Spurfies can synthesize novel views in higher quality than previous sparse-view methods.}
    \label{fig:mip-extra-nvs}
\end{figure*}

\begin{lstlisting}
voxel_size = (0.025, 0.025, 0.025)          % Voxel size for each dimension
voxel_scale = (2, 2, 2)                     % Voxel scale for each dimension
kernel_size = (3, 3, 3)                     % Range of voxels searched for neighbors
max_points_per_voxel = 26                   % Maximum number of points stored in a voxel
max_occ_voxels_per_example = 20000          % Maximum number of occupied voxels per point cloud
ranges = (-1.0, -1.0, -1.0, 1.0, 1.0, 1.0)  % Maximum ranges the VoxelGrid spans
\end{lstlisting}

The $\text{voxel size}$ is set to match the average distance between neural points, ensuring an appropriate spatial distribution. Each voxel is limited to containing a maximum of 26 points, balancing between spatial resolution and computational efficiency.

We set K = 8 for neighbor queries, aiming to have, on average, 8 queried neural points for every ray-marched query point. This configuration strikes a balance between capturing sufficient local information and maintaining computational efficiency.

\paragraph{Datasets:} For evaluation on the DTU~\cite{dtu} dataset, we adhere to the split  established by S-VolSDF~\cite{s-volsdf}. This protocol excludes scans from the training set of multi-view stereo methods, utilizing only those in the test/validation splits. Additionally, we follow the standard protocol employed by~\cite{large, volsdf, regnerf} for the quantitative evaluation.

Since, there are no previous sparse view method tested on Mip-NeRF 360~\cite{mipnerf}, we randomly select three input views for all qualitative evaluations. The lack of ground-truth point clouds precludes Chamfer Distance (CD) evaluation, limiting our analysis to qualitative results. Our evaluation encompasses four scenes from Mip-NeRF 360, with corresponding view IDs as follows:
Garden: (DSC08116, DSC08121, DSC08140)
Kitchen: (DSCF0683, DSCF0700, DSCF0716)
Treehill: (\_DSC9004, \_DSC9005, \_DSC9006)
Stump: (\_DSC9307, \_DSC9313, \_DSC9328). 

\paragraph{Results:}
Our method demonstrates superior performance in mesh reconstruction compared to point-based techniques such as Points2Surf~\cite{points2surf} and CAP-UDF~\cite{cap-udf}. As shown in Figure~\ref{fig:pts_recon}, even when using points obtained from Dust3R~\cite{dust3r}, these alternative methods often produce reconstructions with holes and fail to capture fine details. In contrast, our approach achieves more complete and detailed reconstructions, highlighting the effectiveness of our neural point-based representation and learned local geometry prior. 

Additional reconstruction and Novel View Synthesis (NVS) results on the Mip-NeRF 360 dataset are presented in Figure~\ref{fig:mip-extra} in comparison with NeuSurf~\cite{neusurf} and S-VolSDF~\cite{s-volsdf}. We also show points sampled from the reconstructed mesh Figure~\ref{fig:pc-vis}.

\section{Local Prior}

\paragraph{Data:} To train our local prior, we design a setup that emulates volume rendering conditions. We sample two distinct sets of points:

1) \textbf{Query points} $\mathcal{X} = \left\{ (\mathrm{x}_i,  \mathrm{\rho}_i) \right\}_{i=1}^{\mathrm{N}}$, sampled in close proximity to the mesh surface. These points are generated using two different variances (0.05 and 0.001) to simulate the ray-marching process in volume rendering. On average, we sample $\mathrm{N} = 500k$ query points for every mesh.

2) \textbf{Neural points} $\mathcal{N} = \{(\mathrm{p}_j, \mathrm{f}^g_j)\}_{j=1}^{\mathrm{M}}$, which represent the underlying structure of our reconstruction. To ensure density-agnostic learning during local prior training, we employ farthest point sampling on the mesh surface, maintaining an average inter-point distance of 0.025. The number of neural points, $\mathrm{M}$, varies based on the mesh size, which is normalized to fit within a unit cube. During inference for sparse view reconstruction, we subsample the neural points to match the same density as training.

Our training data is from five classes in the ShapeNet~\cite{shapenet} dataset: sofas, chairs, planes, tables, and lamps. To enhance robustness against noise, we add Gaussian noise with a variance of 0.01 to the neural points. 

\paragraph{Training:} To train the local prior, we employ a combination of loss functions:

\begin{equation}
    \mathcal{L}_{\mathrm{prior}} = \mathcal{L}_{\mathrm{SDF}} + \lambda_{\mathrm{TV}} \cdot \mathcal{L}_{\mathrm{TV}} + \lambda_{\mathrm{eik}} \cdot \mathcal{L}_{\mathrm{eik}}
\end{equation}

where $\lambda_{\mathrm{TV}} = 1\mathrm{e}{-2}$ and $\lambda_{\mathrm{eik}} = 1\mathrm{e}{-3}$. 

Our training process utilizes a batch size of 5. Each batch instance comprises 40,000 randomly sampled query points, equally distributed between positive and negative SDF samples, along with 2,000 neural points. These neural points are padded with points outside the unit cube to ensure consistent batch size.

We train the geometry MLP and the latent codes for 5,000 epochs. For the latent codes, we implement a cosine annealing learning rate schedule, starting at $1\mathrm{e}{-2}$ and gradually decreasing to $3\mathrm{e}{-4}$. The MLP is trained with a constant learning rate of $3\mathrm{e}{-4}$. We use the Adam~\cite{adam} optimizer throughout the training process. Instead of $K=8$ during sparse view reconstruction, we set $K=4$ neighbors during the local prior training.

The total training time for the local prior is approximately 8 hours, utilizing a single A100 GPU. This comprehensive training approach ensures that our local prior effectively captures the geometric properties of diverse shapes, enabling robust sparse-view reconstruction.

\paragraph{Results:}

\begin{figure}[!t]
    \centering
    \includegraphics[width=0.4\textwidth]{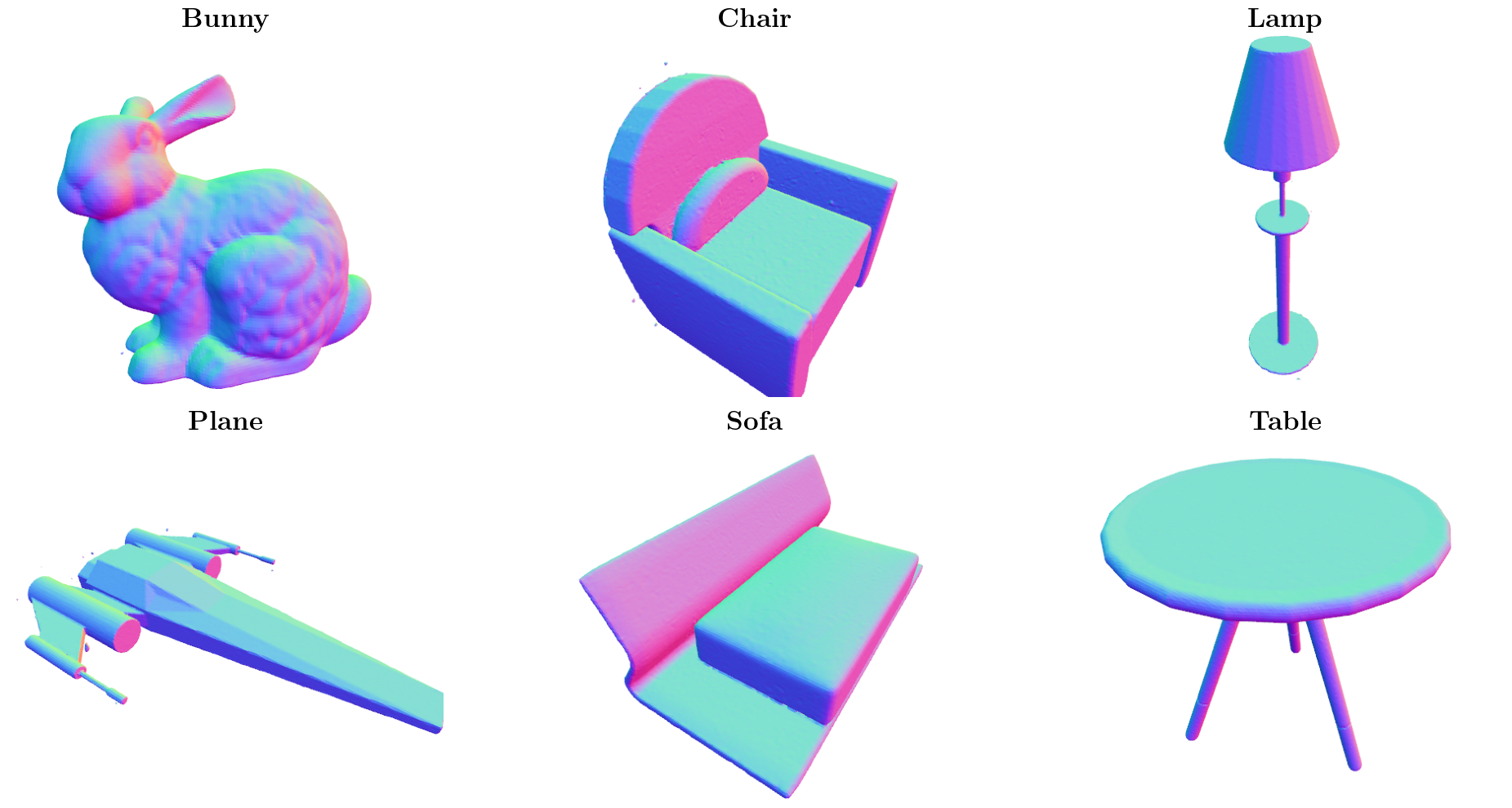}
    \caption{\textbf{Mesh reconstruction} results on a few unseen objects from ShapeNet~\cite{shapenet} and the Stanford bunny.}
    \label{fig:shapnet}
\end{figure}

We show some quantitative results in Figure~\ref{fig:shapnet} of surface reconstruction on ShapeNet~\cite{shapenet} dataset and the Stanford bunny. These results are shown for unseen objects after training the prior. The geometry MLP is frozen and only the geometry latent codes are optimized. We achieve quality mesh reconstruction with high surface details.

\begin{figure}[!t]
    \centering
    \includegraphics[width=0.5\textwidth]{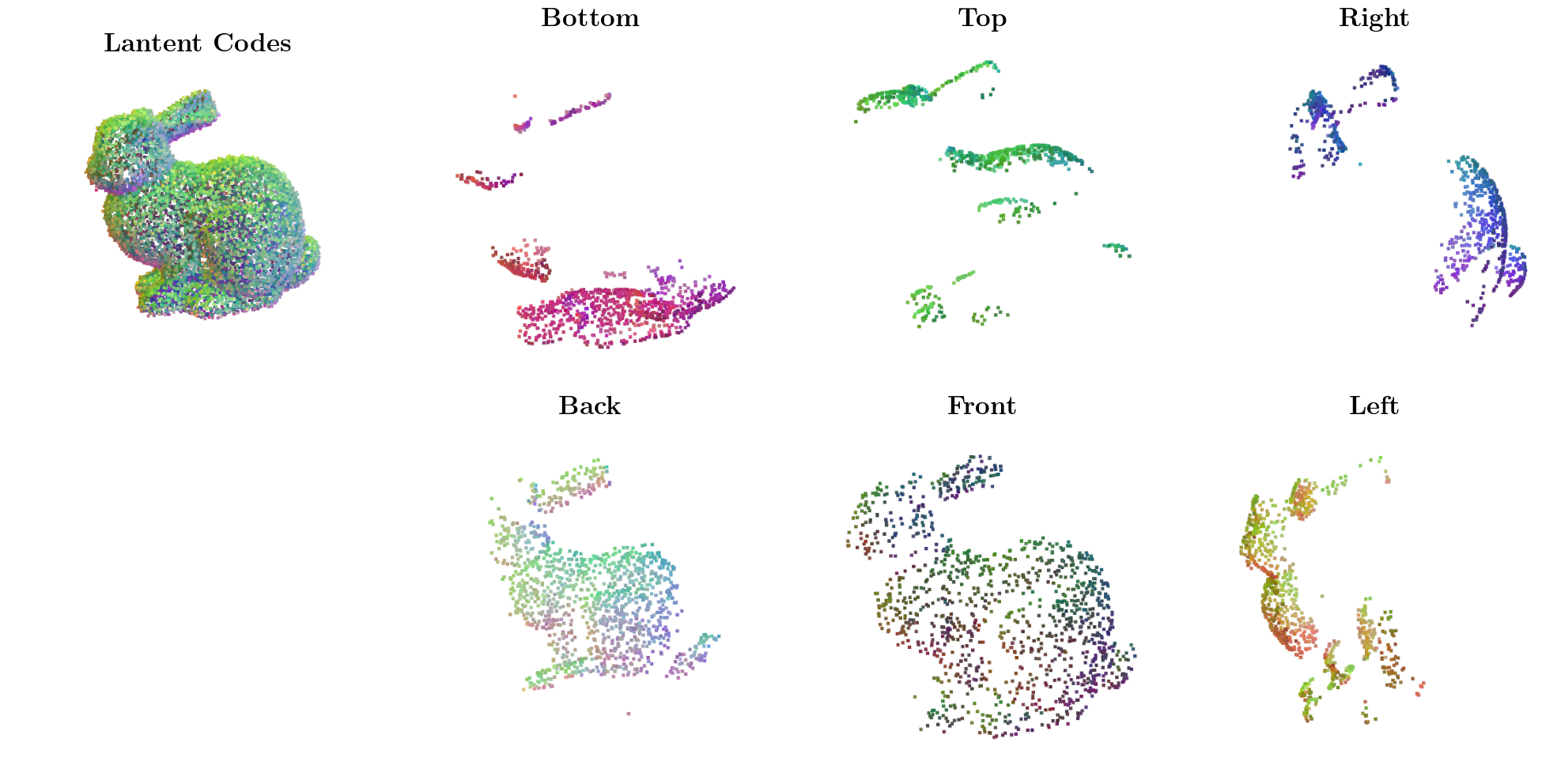}
    \caption{Clustering of optimized geometry latent codes based on six orientations. The geometry latent codes add local descriptive information to point clouds.}
    \label{fig:cluster}
\end{figure}

We extend our analysis to demonstrate the potential of our optimized geometry latent codes for point cloud clustering. Figure~\ref{fig:cluster} illustrates this capability using the Stanford bunny model, where we present six distinct clusters derived from these codes.

The clustering results reveal an property of our optimized geometry latent codes. These codes appear to capture and encode local surface orientations effectively. As a result, the clustering process groups points with similar local geometric characteristics together. This suggests that our method not only reconstructs the surface accurately but also learns a meaningful representation of local surface properties. Specifically, we observe that: 1) Points belonging to the same cluster tend to have similar surface normals or curvature properties.
2) Transitions between clusters are generally smooth, indicating a continuous representation of geometric features which is achieved using $\mathcal{L}_{\mathrm{TV}}$.

This clustering capability demonstrates an additional utility of our approach beyond surface reconstruction. It suggests potential applications in shape analysis, feature detection, and semantic segmentation of 3D models.

\end{document}